\documentclass[preprint,authoryear,12pt]{elsarticle}

\usepackage{amsmath, graphicx, mathtools, amssymb}
\usepackage{multirow}
\usepackage{caption}
\usepackage{subcaption}
\usepackage{url}
\usepackage{booktabs, hyperref}
\usepackage{gensymb}
\usepackage{xcolor, lineno}
\usepackage{multicol}
\usepackage{subcaption}

\definecolor{myblue}{rgb}{0,0,0.75}
\definecolor{mygreen}{rgb}{0,0,0}
\definecolor{myred}{rgb}{0.5,0,0}
\definecolor{hidden}{rgb}{1,1,1}

\newcommand{\Changes}[1]{\textcolor{mygreen}{#1}}

\newenvironment{theorem*}[1]{{\textbf Theorem #1} \begin{itshape}}{\end{itshape}}

\newenvironment{corollary*}[1]{{\textbf Corollary #1} \begin{itshape}}{\end{itshape}}

\newenvironment{proposition*}[1]{{\textbf Proposition #1} \begin{itshape}}{\end{itshape}}

\journal{International Journal of Production Research}

\begin{document}

\begin{frontmatter}

\title{Why do zeroes happen? A model-based approach for demand classification}

\author[label1]{Ivan Svetunkov\corref{cor1}} \ead{i.svetunkov@lancaster.ac.uk}
\author[label1]{Anna Sroginis}
\cortext[cor1]{Correspondence: Ivan Svetunkov, Department of Management Science, Lancaster University Management School, Lancaster, Lancashire, LA1 4YX, UK.}
\address[label1]{Centre for Marketing Analytics and Forecasting \\ Lancaster University Management School, Lancaster, LA1 4YX, UK}

\begin{abstract}
Effective demand forecasting is critical for inventory management, production planning, and decision making across industries. Selecting the appropriate model and suitable features to efficiently capture patterns in the data is one of the main challenges in demand forecasting. In reality, this becomes even more complicated when the recorded sales have zeroes, which can happen naturally or due to some anomalies, such as stockouts and recording errors. Mistreating the zeroes can lead to the application of inappropriate forecasting methods, and thus leading to poor decision making. Furthermore, the demand itself can have different fundamental characteristics, and being able to distinguish one type from another might bring substantial benefits in terms of accuracy and thus decision making. We propose a two-stage model-based classification framework that in the first step, identifies artificially occurring zeroes, and in the second, classifies demand to one of the possible types: regular/intermittent, intermittent smooth/lumpy, fractional/count. The framework relies on statistical modelling and information criteria. We argue that different types of demand need different features, and show empirically that they tend to increase the accuracy of the forecasting methods and reduce inventory costs compared to those applied directly to the dataset without the generated features and the two-stage framework. %Our general practical recommendation based on that is to use the mixture approach for intermittent demand, capturing the demand sizes and demand probability separately, as it seems to improve the accuracy of different forecasting approaches.
\end{abstract}

\begin{keyword}
Intermittent Demand \sep Stockout \sep Classification \sep Forecasting
%% keywords here, in the form: keyword \sep keyword

%% MSC codes here, in the form: \MSC code \sep code
%% or \MSC[2008] code \sep code (2000 is the default)

\end{keyword}

\end{frontmatter}

%%
%% Start line numbering here if you want
%%
% \linenumberscouldn

%%%%%%%%%%%%%%% Introduction %%%%%%%%%%%%%%%
\section{Introduction} \label{sec:Introduction}
People working in the area of demand forecasting sometimes encounter zeroes in their data. These zeroes can happen for a variety of reasons: some of them occur naturally, because no one buys the product in that specific moment of time, some happen artificially due to problems in inventory management system or recording errors. It is important to distinguish these two situations and treat them differently. In case of the naturally occurring zeroes, we typically have intermittent demand, which according to the definition of \cite{Svetunkov2023b} is the demand that ``has non-zero values occurring at irregular frequency''. Many statistical methods have been developed, starting from the \cite{Croston1972} and its bias corrected form by \cite{Syntetos2001} to machine learning methods developed over the years \citep{Hasni2018, Babai2020, Jiang2021a}, so an analyst can select the most appropriate or the favourite approach and use it for intermittent demand forecasting. However, if we deal with the artificially occurring zeroes, they need to be treated differently: for example, using an intermittent demand approach on the data with stockouts would be harmful for decision making, because we would be forecasting stockouts instead of demand.

Furthermore, even when the stockouts are taken into account, it is not clear how to distinguish the intermittent demand from the regular one. The literature has not answered the question ``how many zeroes do you need to have to decide that you deal with intermittent demand?''. And overall, the question ``Why do zeroes happen?'' has been neglected.

Finally, we argue that there can be different types of demand, and using some important features for them can potentially improve the accuracy of the forecasting approaches applied to them. For example, treating the intermittent demand in the same way as the regular one might lead to less accurate point forecasts, which in turn would lead to inefficient decisions.

In this paper, we want to close several gaps in the literature by:

\begin{enumerate}
    \item developing an approach for automated demand classification,
    \item developing an approach to make automatic detection of potential stockouts\footnote{We use the term ``stockout'' to denote any situations with artificially occurring zeroes in the data. This is discussed in detail in Section \ref{sec:TheMethod}.},
    \item suggesting several fundamental features based on (1) and (2) that, as we argue, should improve the performance of approaches in terms of forecasting and inventory management.
\end{enumerate}

%%%%%%%%%%%%%%% Literature review %%%%%%%%%%%%%%%
\section{Literature review} \label{sec:litReview}
\subsection{What is ``intermittent demand''?}
The rise of the interest in the area of intermittent demand started with the paper of \cite{Croston1972}, who acknowledged that the simple exponential smoothing \citep[SES by][]{Brown1956} is not appropriate when used on the data with unpredictable zeroes. To solve the problem, he suggested to split the data into two time series: demand sizes and demand intervals. His idea was that the demand that we observe ($y_t$) can be represented as a combination of two variables:
\begin{equation} \label{eq:CrostonIdea}
    y_t = o_t z_t,
\end{equation}
where $z_t$ is the demand size on observation $t$ and $o_t$ is a binary variable of demand occurrence, that has a probability of occurrence $p_t = \frac{1}{q_t}$, where $q_t$ is the interval between the observed demand sizes. To produce forecasts, \cite{Croston1972} used two simple exponential smoothing methods for capturing dynamics of the demand sizes and demand intervals. While being efficient and innovative, his approach was neglected by academia for more than 20 years until \cite{Willemain1994} and \cite{Johnston1996} showed that Croston's method performed well in practice and should be preferred for intermittent demand instead of other simple forecasting methods. Acknowledging the existence of intermittent demand, these papers also opened a new direction of research -- intermittent demand forecasting, where the patterns of the data are so different that the conventional forecasting techniques might fail or not work efficiently.

From the practical point of view, when separating the regular demand from the intermittent, one still faces a challenge, because there are no appropriate rules and it is not clear, what quantity of zeroes transforms the regular demand into intermittent. Some practitioners use arbitrary thresholds of 10\%, 15\%, 20\% etc of zeroes in the data as cut-off points, where one would need to switch from regular to intermittent demand forecasting method. But those thresholds do not have any theoretical rationale behind them, and can only be considered as approximations to the real solution of the problem. Even non-OR methods that could be used to forecast special events are still based on some arbitrary threshold values which differ across domains \citep{Nikolopoulos2021}. So far, the assumed ``rule of thumb'' is that any quantity of zeroes implies that the demand is intermittent.

Furthermore, the academic definitions of intermittent demand found in the literature are generally broad. Early definitions, such as \cite{Silver1981}, describe it as demand with ``significant periods of no demand activity'' but the term ``significant'' is unclear. Later definitions, including \cite{Willemain1994} and \cite{Syntetos2001}, describe it as random demand with many (or large proportion of) zero periods, but the terms ``many'' and ``large proportion'' used remain ambiguous. More recent definitions of \cite{Syntetos2005}, \cite{Syntetos2009}, \cite{Teunter2011}, and \cite{Babai2014a} describe intermittent demand as appearing sporadically, which is, in essence, correct, but does not distinguish between random absence of demand and systematic gaps caused by external factors such as disruptions or recording errors. 

Lately, some definitions became more case-specific and sometimes confusing, like ``the features of intermittent demand are characterised by their irregularity, with a very small demand size'' \citep{Jiang2021a}, or that it is ``characterised by time series with many zeroes'' \citep{Prestwich2021}, or characterised by ``irregular demand occurrence and low demand quantity variation'' \citep{Rozanec2022}.

One other mistake sometimes made in the literature is equating count demand to the intermittent one \citep{Snyder2012}. While in some situations this is correct, this is not universally the case because the term ``intermittent'' relates to the demand intervals, while the word ``count'' describes demand sizes. In reality, there are many examples of demand being intermittent yet fractional (e.g. electricity vehicles charging).

Among all definitions, we find the one by \cite{Svetunkov2023b} to be the most concise and clear: ``\textbf{intermittent time series is a series that has non-zero values occurring at irregular frequency}''. This definition shows that the non-zero values happen randomly and cannot be predicted, and the definition does not impose any subjective terms like ``many'', or ``large'', or ``some''. However, it still does not make a distinction between naturally occurring zeroes and the ones happening due to external factors.

We argue that it is extremely important to distinguish between the two types of zeroes, especially in retail settings, since many companies struggle to detect correctly stock levels due to stockouts, and product unavailability problems. We propose differentiating between pure randomness and explainable absences by attempting to capture anomalies in the data using an automatic detection tool for potential stockout situations (more on this in Section 2.4).

%%%%% Demand classification %%%%%
\subsection{Demand classification}
There are several proposed classification schemes based either on variance partition, or the accuracy of forecasting procedures \citep{Babiloni2010}. We discuss the main ones here.

% \subsubsection{Classification schemes based on variance partition}
\cite{Williams1984} was one of the first who proposed classifying demand into `smooth', `slow-moving' and `sporadic' by partitioning the variance of demand over the lead time into causal parts. \cite{Eaves2004} expanded this classification by inclusion of the irregular type, which is differentiated from the smooth one according to the level of demand sizes variability.
% (some `smooth' patterns in \cite{Williams1984}, mildly intermittent and highly intermittent (as a segregation of `sporadic')).
However, both papers clearly state that the boundaries between these categories are industry/data-specific, and they should come as a managerial decision.

% \subsubsection{Classification schemes based on the forecast accuracy}
\cite{Syntetos2005a} proposed a classification (now called `SBC', Syntetos-Boylan-Croston) based on other principles. They discussed the existing categorisation schemes and compared Croston, Simple Exponential Smoothing (SES) and Syntetos-Boylan Approximation \citep[SBA from][]{Syntetos2005} based on their theoretical Mean Square Error (MSE) values. The authors showed that one can derive the cut-off values for average demand intervals and coefficient of variation, allowing to select between Croston and SBA for each type of demand. This scheme has gained large popularity among practitioners, because it suggest four distinct categories of intermittent demand:

\begin{enumerate}
    \item Smooth;
    \item Intermittent but not very erratic;
    \item Lumpy;
    \item Erratic but not very intermittent.
\end{enumerate}

In many cases practitioners use the scheme as a prior step for data analysis, categorising the intermittent demand into these categories without any specific purpose, completely neglecting it in the following steps, where, for example, they use some machine learning methods for demand forecasting. We should point out that the original motivation of \cite{Syntetos2005a} was to help in choosing between Croston and SBA, rather than doing a categorisation for the sake of it. This idea seems to have been lost over the years.

The modification of the SBC was proposed by \cite{Kostenko2006}, who modified it, using an inequality from the ratio of MSEs from the original paper of \cite{Syntetos2005a}. They demonstrated that the parameter space for different forecasting methods has a non-linear cut-off. This classification, while being mathematically correct, has not gained as much popularity as SBC, being more complex.

Finally, \cite{Petropoulos2015} proposed a further refinement of the \cite{Kostenko2006} scheme by adding SES method to the classification. They showed on the example of the Royal Airforce data (RAF) that this addition produces more accurate and less biased forecasts than the \cite{Kostenko2006} scheme.

However, all the research that has been done in this direction up to this point has mainly focused on selecting between Croston and SBA specifically for intermittent demand. This implies that if there is at least one zero in the data, then the data can be flagged as intermittent and either of these two methods should be used. While this is widely true, there is evidence from \cite{Syntetos2006} that SES and Simple Moving Average (SMA) perform well even on data with some zeroes. So a refined approach is needed in order to classify demand as intermittent or regular.

%%%%% Intermittent demand approaches %%%%%
\subsection{Intermittent demand approaches}
If we aim to develop a practical classification scheme, we must first understand what kind of models are typically applied in the context of intermittent demand. We do not aim to discuss all literature in the area -- this would be a futile task, given the number of papers. Neither do we aim to find the best forecasting model for intermittent demand. Rather, we aim to use an approach to showcase the potential benefits of our classification scheme. Therefore, in this subsection, we briefly introduce the most popular forecasting methods and models for intermittent time series: (1) Statistical methods, including exponential smoothing; (2) Combination approaches; (3) Machine learning methods developed for intermittent demand.

%%%%% Exponential smoothing %%%%%
\subsubsection{Exponential smoothing}

Research on exponential smoothing for intermittent demand has advanced from early explorations of Croston’s method and its statistical foundations \citep{Shenstone2005} to more refined models tailored to count and intermittent data. \cite{Hyndman2008b} introduced the Hurdle Shifted Poisson filter, aligning with Croston’s forecasts while accounting for count data. Building on this, \cite{Snyder2012} and \cite{Taylor2012} proposed Poisson- and Negative Binomial-based filters, with the latter offering improved forecast accuracy. Moving beyond integer demand assumptions, \cite{Teunter2011} developed a dual SES approach forecasting demand size and occurrence probability directly, which has been used since then in a wider variety of contexts \citep[for example, in][]{Babai2014a, Segerstedt2023, Doszyn2024}. Most recently, \cite{Svetunkov2023b} introduced an ETS-based model combining Bernoulli and positive distributions (e.g., Gamma), demonstrating superior forecasting performance for both fractional and count data.

Remarkably, all the approaches above split the intermittent demand into two parts (demand sizes and demand intervals/demand occurrence) and seem to gain in accuracy by doing so.

%%%%% Combinations %%%%%
\subsubsection{Combination approaches}

Another key research stream in intermittent demand focuses on aggregation and combination approaches. \cite{Nikolopoulos2011a} proposed aggregating intermittent demand to a regular demand level, forecasting it, and then disaggregating back to the original level, allowing the use of conventional forecasting methods without the common challenges of intermittent data. \cite{Petropoulos2015} extended this idea by combining forecasts across different aggregation levels. \cite{Kourentzes2021} applied temporal hierarchies to combine point forecasts, capturing hidden structures like trends and seasonality in intermittent demand. Finally, \cite{Wang2024} introduced probabilistic forecast combinations, showing that simple average combination performs best for quantile forecasts and inventory metrics.

%%%%% Machine learning %%%%%
\subsubsection{Machine learning for intermittent demand}
The paper by \cite{Kourentzes2013} was the first one that we are aware of that used Artificial Neural Networks (ANN) for intermittent demand forecasting. He proposed two architectures: one capturing demand sizes and intervals in Croston’s style before dividing them, and another directly forecasting final demand. Both used demand sizes and intervals as inputs. While these methods did not outperform simpler forecasting approaches (e.g., Croston's method), they showed slight improvements in inventory metrics.

\Changes{\cite{Nikolopoulos2016} evaluated performance of the k-nearest neighbour method for intermittent demand forecasting, comparing it with the conventional forecasting methods, showing that it outperforms Croston, SBA and SES on real data.}

\cite{Babai2020} compared SES, Croston, SBA, and bootstrapping methods with a Multilayer Perceptron ANN on a spare parts dataset, showing that a well-designed neural network could outperform conventional methods in point forecasts and inventory costs. Similarly, \cite{Turkmen2019} used a recurrent neural network with Negative Binomial and Poisson distributions for demand sizes and parametric distributions for demand intervals. Their approach, tested on several intermittent demand datasets, proved competitive with statistical methods, sometimes outperforming them in point forecasts and specific quantiles.

\cite{Jiang2021a} introduced an adaptive Support Vector Machine for spare parts demand forecasting, comparing it with parametric, bootstrap, and neural network methods. Their approach performed well in Mean Absolute Error (MAE) and scaled Mean Error, but the chosen error measures, minimised by the median, which in intermittent demand can often lead toward models predicting values closer to zero.

\cite{Rozanec2022} proposed separating demand into regular and intermittent categories, using a gradient-boosted decision tree (CatBoost) to predict demand occurrence and a light gradient boosting machine (LightGBM) for demand sizes. Their approach outperformed conventional methods (e.g., Naïve, SES, SMA) in Area Under the Curve (AUC) and Mean Absolute Scaled Error (MASE). However, their methodology had drawbacks: (1) MASE is minimised by median, potentially selecting models biased toward zero demand; (2) instead of using probability of occurrence, the authors applied an arbitrary threshold to classify forecasts as zero or one. While suitable for classification, this is problematic for intermittent demand, because the occurrence of intermittent demand is fundamentally unpredictable. By forcing a binary classification, the model risked capturing noise rather than underlying patterns. Nonetheless, the idea of distinguishing between regular and intermittent demand remains valuable.

\cite{Shrivastava2023} developed few Recurrent Neural Networks that would produce forecasts for demand sizes and probability of occurrence and demonstrated that this approach outperforms Croston, SBA and TSB on the M5 and car parts datasets.

\Changes{In a recent paper \cite{Zhang2024} use Transformer architecture to forecast intermittent demand. While the idea is interesting, the main flaw of the paper is in the evaluation, where the authors violate the basic forecast evaluation principles (no explanation of the forecast horizon, inadequate error measures used etc).}

Many other studies have explored machine learning for intermittent demand. While we do not aim to cover all of them, we want to note that approaches that separately model demand sizes and demand occurrence tend to perform well. Additionally, it appears beneficial to apply different forecasting methods depending on the data type, rather than using a single model for all cases. We aim to use some of these findings in our experiments and case studies.

%%%%% Stockouts %%%%%
\subsection{Stockout/Out-of-Stock identification}
As mentioned in the introduction, zeroes in the data can occur either naturally or artificially. In the former case, we would be talking about the canonical intermittent demand, where zeroes represent the situation when nobody buys our product. In the latter case, zeroes can occur for a variety of reasons, such as (i) stockouts; (ii) absence of product on shelves; (iii) product being (temporary) discontinued; (iv) no sales due to calendar events (e.g. shop closed during Christmas holidays); (v) recording errors and others.

\Changes{We note that our paper does not focus on the censored demand and its implications to the inventory decisions, although there are many papers on the topic \citep[see, for example,][]{Lau1996, Sachs2014, Trapero2024}. Instead, we focus on the demand classification, arguing that stockouts detection is one of the first important steps an analyst needs to make.}

Generally, the retail operations literature has focused on the customer reactions to an Out-of-Stock (OOS) situation from the marketing perspective \citep[e.g.][]{Campo2000, Verbeke1998}, while some studies look at the extent and root cause analysis of stockout situations, mainly connecting these to either retail store replenishment causes or upstream problems \citep{Aastrup2010}. There seem to be two main methods for auditing OOS/stockout situations in practice using: (1) shelves images/scanning \citep[e.g.][]{iki2024, Rosado2016} or (2) a data-driven approach based on point-of-sales (POS) data. Although both methods might be expensive and require additional tools and understanding, we would argue that the latter is easier to implement for most cases. 

\cite{fisher2010} advocate for a data-driven analytical approach to improve retail supply chain performance by leveraging customer transaction data, demand forecasting, and inventory optimization techniques. They proposed to use dynamic inventory management, where retailers adjust stock levels based on demand patterns, seasonal fluctuations, profits and store-specific trends. Clearly, there is a need for automatic or semi-automatic detection of stockout via any available methods.

\cite{Papakiriakopoulos2009} proposed a rule-based decision support for the detection of OOS products based on heuristic rules. Their method analyses POS data, inventory records, and historical sales trends, applying predefined rules to identify anomalies -- such as sudden sales drops or discrepancies between stock levels and sales activity, that may indicate stockouts. Using an iterative process of physical audit and classification models with internal and external validation, the authors were able to detect about one third of the OOS cases accurately. The authors note that while the system demonstrates acceptable levels of predictive accuracy and problem coverage, it may not account for all variables influencing shelf stock levels, such as sudden changes in consumer behaviour or external factors affecting sales patterns. \cite{Chuang2018} expanded these ideas by including cost factors in the modelling.

Finally, \cite{Fildes2022} presented forecasting research in presence of stockouts in retail setting. The stockouts themselves can be complete or partial. In case of the former, the product is unavailable and we record zero sales. The latter implies that we run out of product in the middle of the day, so we cannot satisfy the whole demand. In this paper, we focus only on the complete stockouts, because the partial ones can only be identified if the stock system records the data correctly -- it is not always possible to identify the partial stockouts correctly just by analysing sales.

While we acknowledge that there can be many reasons for artificially occurring zeroes, for the purposes of this paper, we call all of them ``\textit{stockouts}''. Besides, in the literature, the terms stockouts, out-of-stock, stock shortage, or out-of-shelf are typically used interchangeably.

In this paper, we use a data-driven approach on point-of-sales data to identify any anomalies in the data that could be associated with stockouts. However, our approach is much simpler than the approaches mentioned above, and it can be potentially substituted by the more advanced ones without changing the essence of the classification approach.

%%%%%%%%%%%%%%% The Method %%%%%%%%%%%%%%%
\section{Model-based Identification Method} \label{sec:TheMethod}
The demand with some stockouts and without any other sources of zeroes can be considered as regular, i.e. the demand that happens on every observation, just with some missing values. In its turn, it can be either count, or fractional: the former has values that take exclusively integer values (not necessarily having zeroes), while the latter is the demand that has fractional values and/or is the demand that has large volume and thus can be modelled using a fractional distribution. In the latter case, when people buy thousands of units of product, some conventional distributions (such as normal) can be used efficiently for modelling and forecasting instead of, for example, Negative Binomial one, which starts behaving like the Normal one on large volume data.

The demand with naturally occurring zeroes can be considered as a proper intermittent. This type of demand can be count or fractional as well, similarly to the regular one. Depending on whether it is a regular or intermittent time series, an analyst can use an appropriate forecasting technique. For instance, if the demand is ``count regular'', a Negative Binomial based model can be used. If it is ``fractional regular'', any conventional forecasting model, such as ETS or ARIMA, or any machine learning technique, can be used. Following \cite{Syntetos2005}, we propose to split intermittent into smooth and lumpy to capture different customers behaviour, but our split comes from the understanding of the possible demand process rather than the minimum of MSE.

Summarising, any demand can be classified into one of the following categories:
\begin{enumerate}
    \item Regular count/fractional;
    \item Intermittent count/fractional:
    \begin{enumerate}
        \item \textit{Smooth intermittent}, where zeroes are considered just a part of the distribution. An example in this category is a product that is sold by a retailer in small quantities every day;
        \item \textit{Lumpy intermittent}, where zeroes have their own dynamics, which can be captured using a separate model. An example here, is a product that is sold occasionally and in bulks.
    \end{enumerate}
\end{enumerate}

\begin{figure}[htb]
    \centering
    \includegraphics[width=\textwidth]{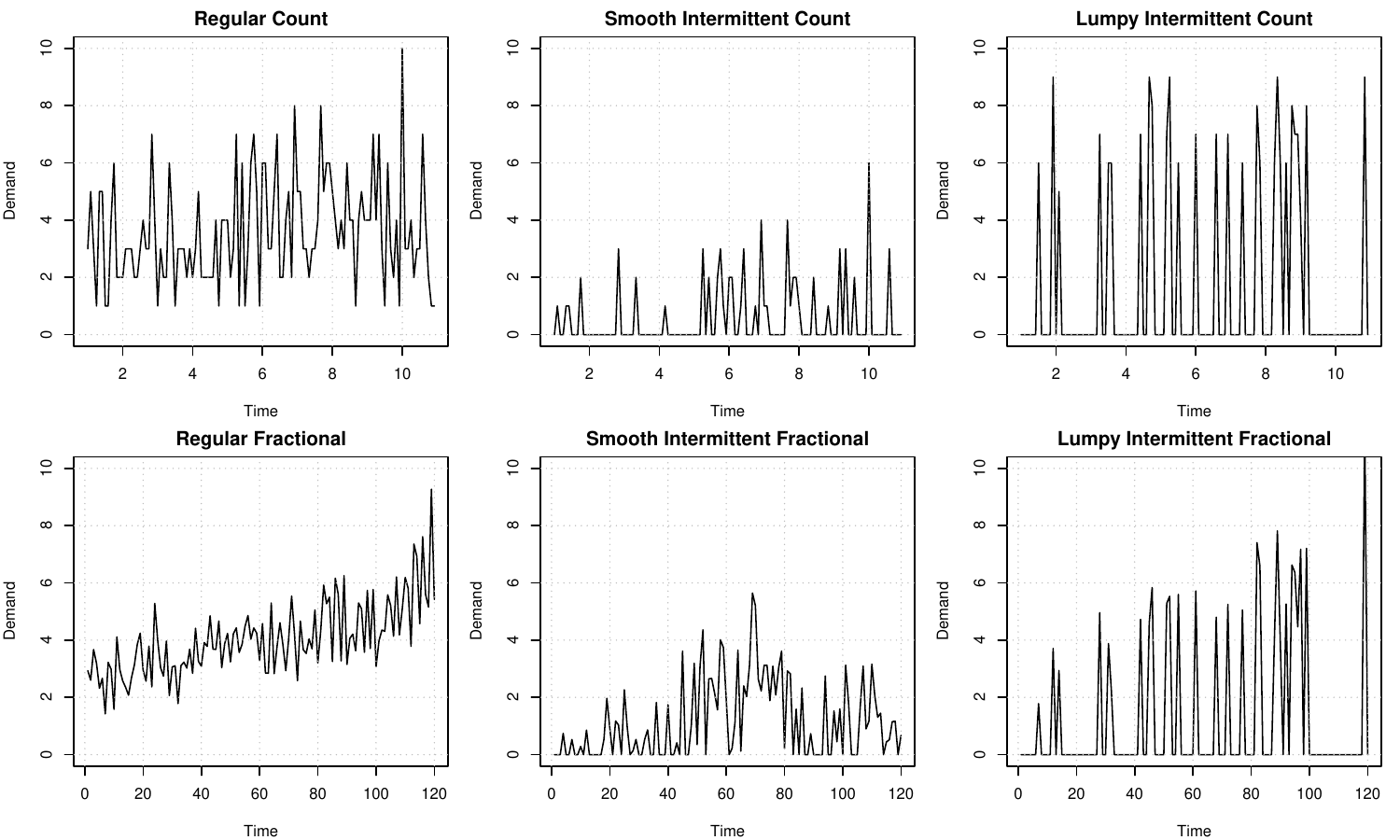}
    \caption{Examples of demand in different categories}
    \label{fig:Categories-Full}
\end{figure}

Figure \ref{fig:Categories-Full} depicts examples of different time series from all the categories mentioned above.

We avoid any arbitrary thresholds for average demand intervals or coefficient of variation because they inevitably assume that the demand occurrence and/or demand sizes do not change over time substantially.

To make such a classification practical, we develop an algorithm relying on several simple statistical models and in-sample selection using information criteria. But before we do this classification, we need to identify and treat the potential stockouts.

%%%%% stockouts %%%%%
\subsection{Identifying stockouts}
To achieve this, we extract the demand intervals $q_{j_t}$ from the data, similar to how they were originally proposed by \cite{Croston1972}, by calculating the difference between the indices of consecutive non-zero observations. This means that two consecutive non-zero demands will have a demand interval of one between them. Assuming that demand occurrence follows a Bernoulli distribution with a time-varying probability $p_t$, the demand intervals should follow the Geometric distribution:
\begin{equation} \label{eq:Geometric}
    q_{j_t}-1 \sim \mathcal{G}(p_t).
\end{equation}
To estimate the time-varying probability, we fit a Friedman's Super Smoother \citep{friedman1984variable} to the series $q_{j_t}$, accounting for potential changes in occurrence probability, for example due to demand becoming obsolete. While other smoothers could be used instead of this one \citep[e.g. LOWESS by][]{Cleveland1979}, we found that the Super Smoother is sensitive enough to capture the potential changes in the demand intervals length. The smoothed series $\hat{q}_{j_t}$ is then used to compute $\hat{p}_t=\frac{1}{\hat{q}_{j_t}}$. Next, we identify observations exceeding a threshold $\nu$, determined by the quantile function of the Geometric distribution. For instance, setting $\nu$ to be equal to 0.99 marks the top 1\% of values as potential stockouts.

\begin{figure}[!htb]
	\centering
    \begin{subfigure}[t]{0.49\textwidth}
	   \centering
    	\includegraphics[width=1\textwidth]{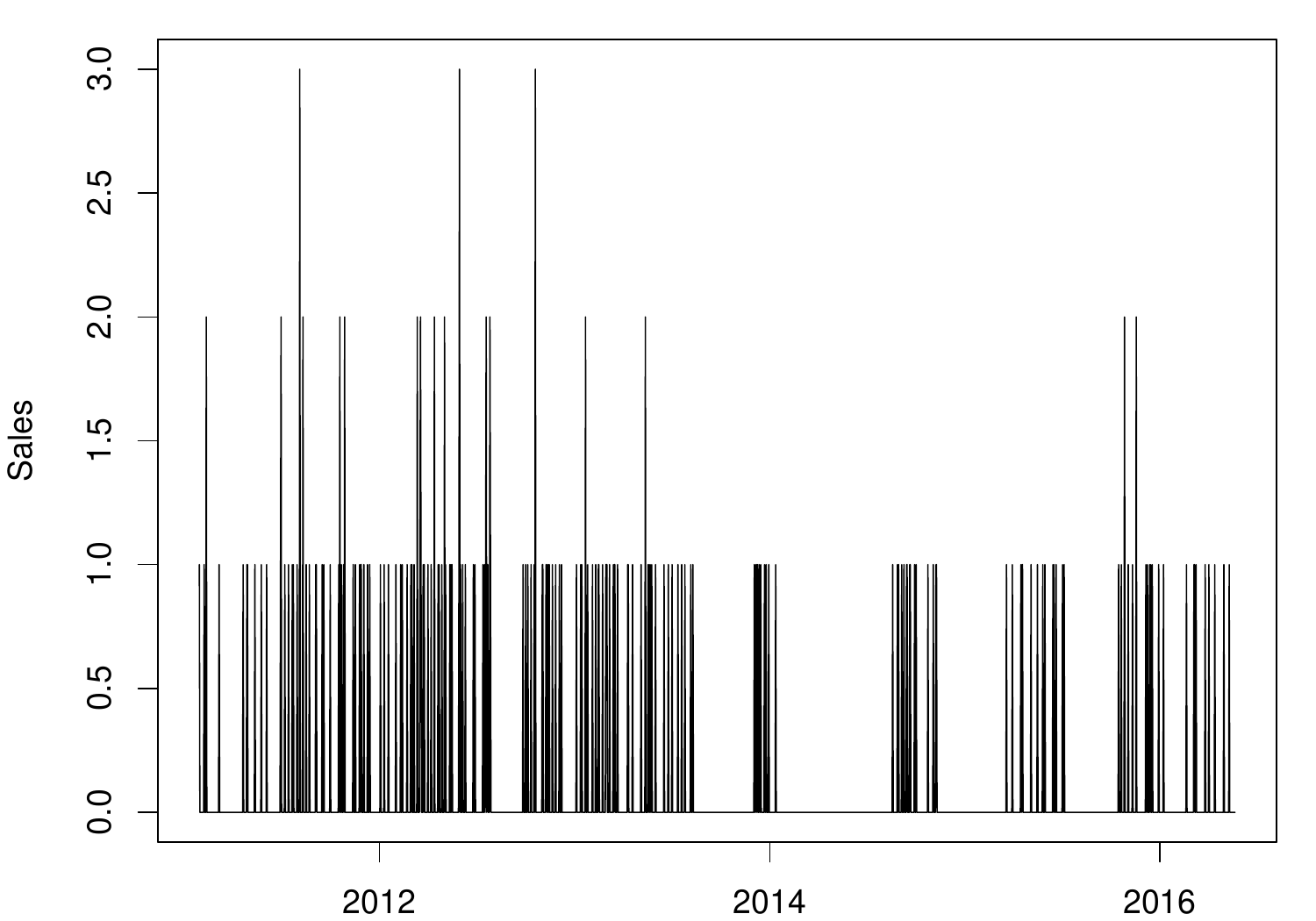}
        \caption{The original time series.}
        \label{fig:M5-Example-original}
    \end{subfigure}
    \begin{subfigure}[t]{0.49\textwidth}
	   \centering
    	\includegraphics[width=1\textwidth]{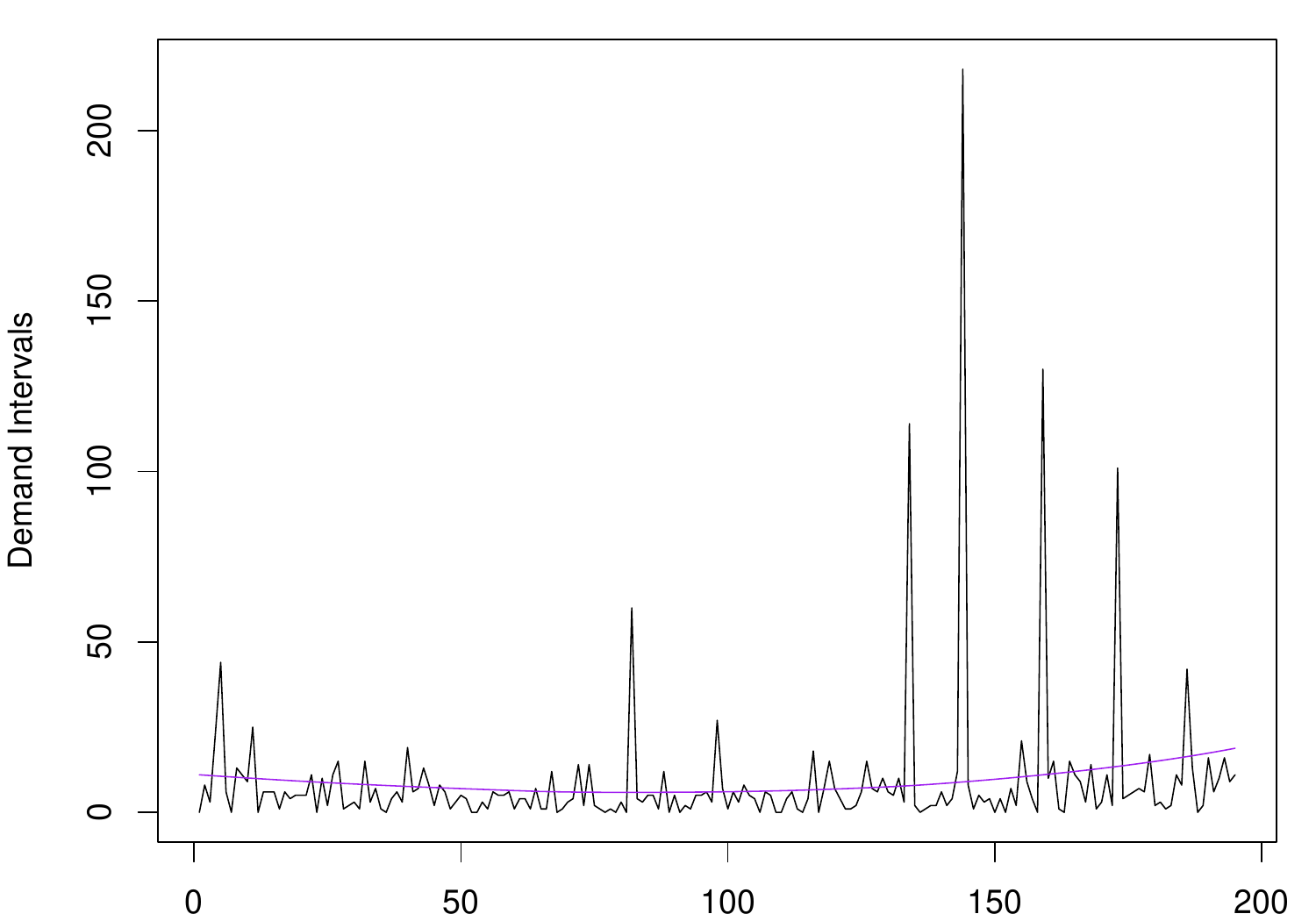}
        \caption{Demand intervals.}
        \label{fig:M5-Example-intervals}
    \end{subfigure}
    \\
    \begin{subfigure}[t]{0.49\textwidth}
	   \centering
    	\includegraphics[width=1\textwidth]{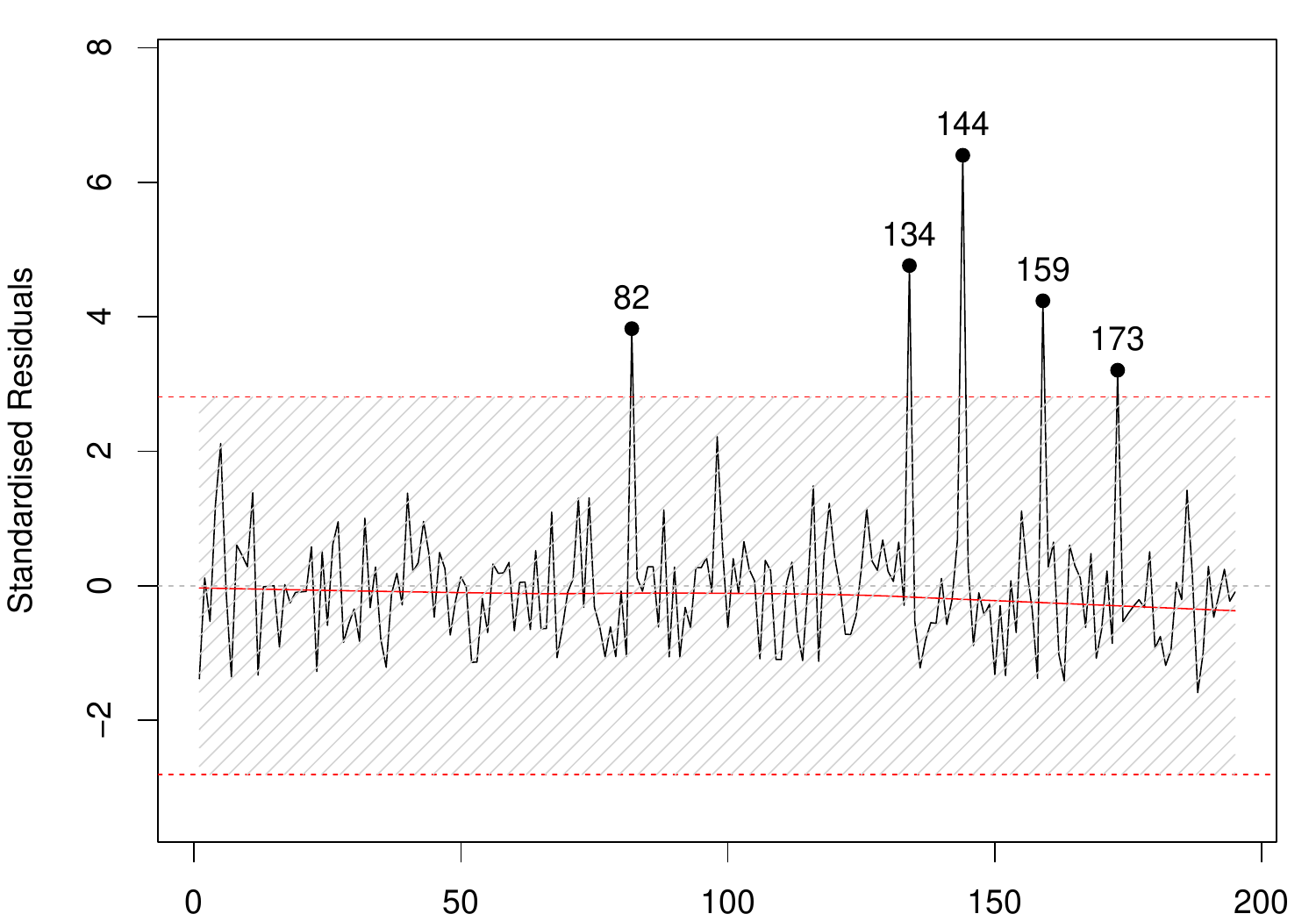}
        \caption{Standardised values.}
        \label{fig:M5-Example-intervals-resid}
    \end{subfigure}
    \begin{subfigure}[t]{0.49\textwidth}
	   \centering
    	\includegraphics[width=1\textwidth]{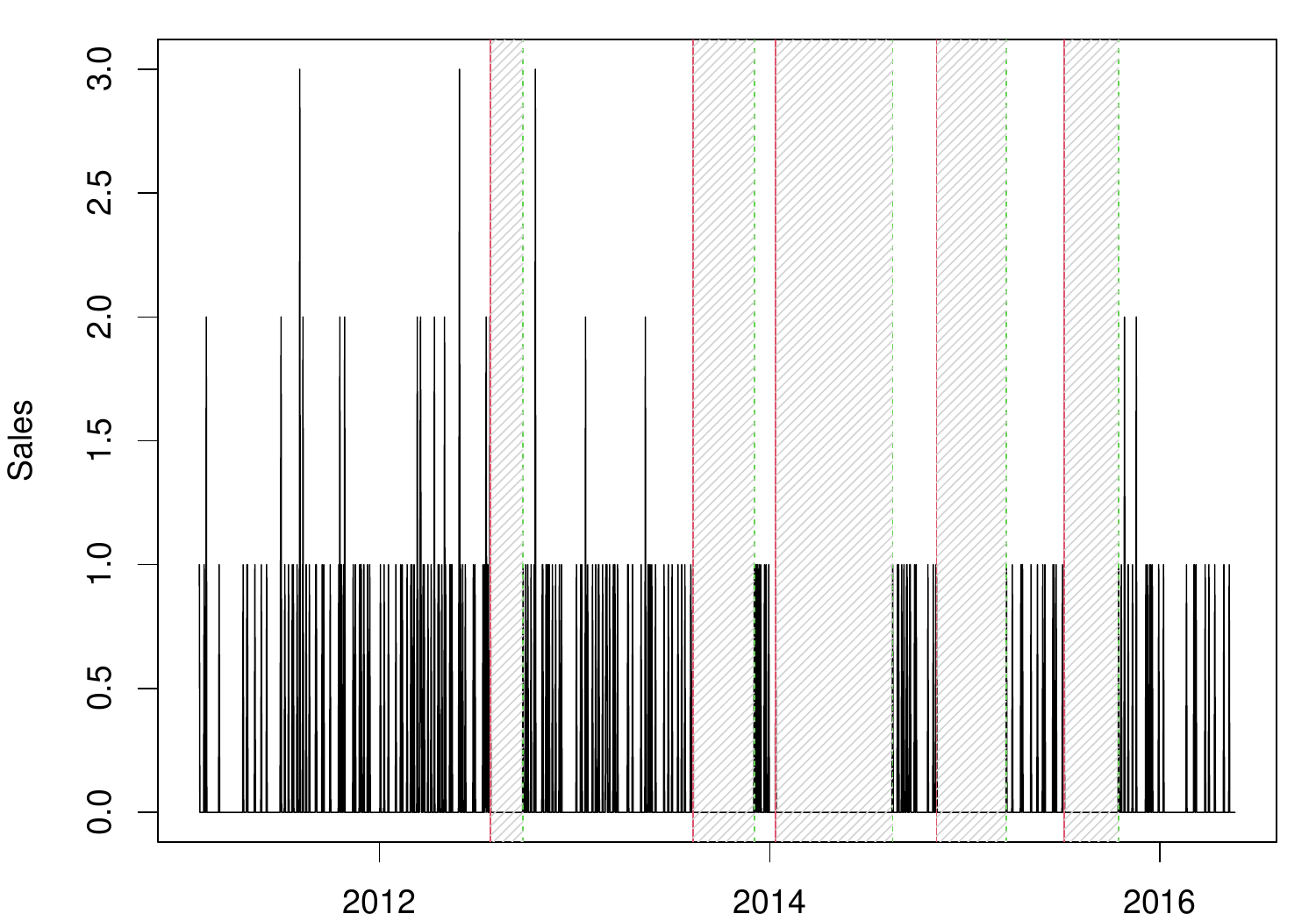}
        \caption{Original series with detected stockouts.}
        \label{fig:M5-Example-stockouts-identified}
    \end{subfigure}
    \caption{Series with stockouts.}
    \label{fig:M5-Example}
\end{figure}

To demonstrate the logic with stockouts identification, we consider an example of an intermittent time series \citep[N10514 from the M5 dataset][]{Makridakis2022}, which is shown in Figure \ref{fig:M5-Example}.

The original time series (Figure \ref{fig:M5-Example-original}) shows several apparent stockouts as zero sales gaps, a few between 2013 and 2016. These stockouts are further highlighted in the demand intervals plot (Figure \ref{fig:M5-Example-intervals}), where they appear as distinct outliers. The solid purple line represents the smoothing applied to detect variations in demand intervals. Using the Geometric distribution model with $\nu=0.999$, we identify several potential stockouts, marked as outliers in Figure \ref{fig:M5-Example-intervals-resid}. Finally, these flagged points are overlaid onto the original time series in Figure \ref{fig:M5-Example-stockouts-identified}, where grey areas indicate detected stockouts. The model successfully identifies the most apparent stockouts.

Given the nature of this method, we argue that stockout identification is influenced by the following factors:

\begin{enumerate}
\item \textit{Number of stockouts}: more stockouts make it harder to distinguish them from naturally occurring zeroes.
\item \textit{Length of stockouts}: longer streaks of zeroes are easier to detect than the shorter ones.
\item \textit{Sample size}: with the same number of stockouts, detection is easier in larger datasets than in the smaller ones.
\end{enumerate}

If a potential stockout occurs at the very first observation, it may indicate that data recording began late, possibly due to a new product. Conversely, if a stockout appears at the end of the series, the product may no longer be sold, or it could represent a recent stockout.

Once stockouts are identified, they need to be addressed. The simplest approach is to remove them from the original data before proceeding to the next step in demand identification. \Changes{We note that we remove the stockout observations purely for the classification purposes to decide whether the demand is indeed intermittent or regular. But we do not recommend doing that for forecasting or inventory management.}

%%%%% Intermittent or not? %%%%%
\subsection{Automated Identification of Demand (AID)}
We propose a model-based method of intermittent demand identification, which we call ``Automated Identification of Demand'', or ``AID''. It relies on the construction of several statistical models and the selection of the most appropriate one using information criteria.

If removing stockouts leaves no zeroes, \textbf{the demand is regular}, and we only need to determine if it is fractional or count. If zeroes remain, \textbf{the demand is intermittent}, and we can determine whether it is lumpy or smooth. Thus AID can be considered as a two-level approach, where the first one (regular/intermittent) is determined automatically after the stockouts are removed, and the second one requires further investigation.

To do the checks for the type of intermittent demand, we propose the following procedure. First, we fit a smooth line \citep[such as Friedman's Super Smoother or LOWESS, respectively by][]{friedman1984variable, Cleveland1979} to the overall demand $y_t$, the demand sizes $z_t$ and to the demand occurrence $o_t$, capturing the potential changes in the dynamics of the data. We thus obtain three smoothed series, $\hat{y}_t$, $\hat{z}_t$ and $\hat{p}_t$ respectively. After that, we use them in fitting several regression models for each of the categories of demand:
\begin{enumerate}[I.]
    \item \label{item:RegularFrac} \textbf{Regular Fractional} -- the model applied to the data itself, $y_t \sim \mathcal{N}(\beta_0 + \beta_1 \hat{y}_t, \sigma_y^2)$\footnote{\Changes{$\mathcal{N}(\mu, \sigma^2)$ denotes the normal distribution with the mean $\mu$ and the variance $\sigma^2$.}}, where $\beta_j$ is a parameter of the model;
    \item \label{item:RegularCount} \textbf{Regular Count} -- Negative Binomial distribution, $y_t \sim \mathcal{NB}(\beta_0 + \beta_1 \hat{y}_t, s_y)$\footnote{\Changes{$\mathcal{NB}(\mu, s)$ denotes the Negative Binomial distribution with the mean $\mu$ and the scale $s$.}}, where $s_y$ is the scale of distribution, estimated together with other parameters of the model. We use this distribution as one of the most flexible count ones;
    \item \label{item:IntermittentSmoothFrac} \textbf{Smooth Intermittent Fractional} -- the model applied to the data itself, $y_t \sim \mathrm{rect}\mathcal{N}(\beta_0 + \beta_1 \hat{y}_t, \sigma_y^2)$ -- this model uses the Rectified Normal distribution, which substitutes negative values with zeroes;
    \item \label{item:IntermittentLumpyFrac} \textbf{Lumpy Intermittent Fractional} -- the mixture distribution model: $y_t = o_t z_t$, where for demand sizes, $z_t \sim \mathcal{N}(\log \beta_0 + \beta_1 \hat{z}_t, \sigma_z^2)$ and for the probability of occurrence, $o_t \sim \mathrm{Bernoulli}(\beta_0 + \beta_1 \hat{p}_t)$\footnote{\Changes{$\mathrm{Bernoulli}(p)$ denotes the Bernoulli distribution with the probability of having one equal to $p$.}};
    \item \label{item:IntermittentSmoothCount} \textbf{Smooth Intermittent Count} -- same as \eqref{item:RegularCount}, but with zeroes. Also, a special case of the {Smooth Intermittent Count} demand is the demand, where only zeroes and a non-zero value occur at random (e.g. when people buy a fixed amount of product), which can  also be called ``\textit{Smooth Intermittent Binary}'' demand;
    \item \label{item:IntermittentLumpyCount} \textbf{Lumpy Intermittent Count} -- the mixture distribution $y_t = o_t z_t$, where demand sizes are $z_t \sim \mathcal{NB}(\beta_0 + \beta_1 \hat{z}_t, s_z)$ and occurrence is $o_t \sim \mathrm{Bernoulli}(\beta_0 + \beta_1 \hat{p}_t)$;
\end{enumerate}

\begin{figure}[!htb]
    \centering
    \includegraphics[width=1\textwidth]{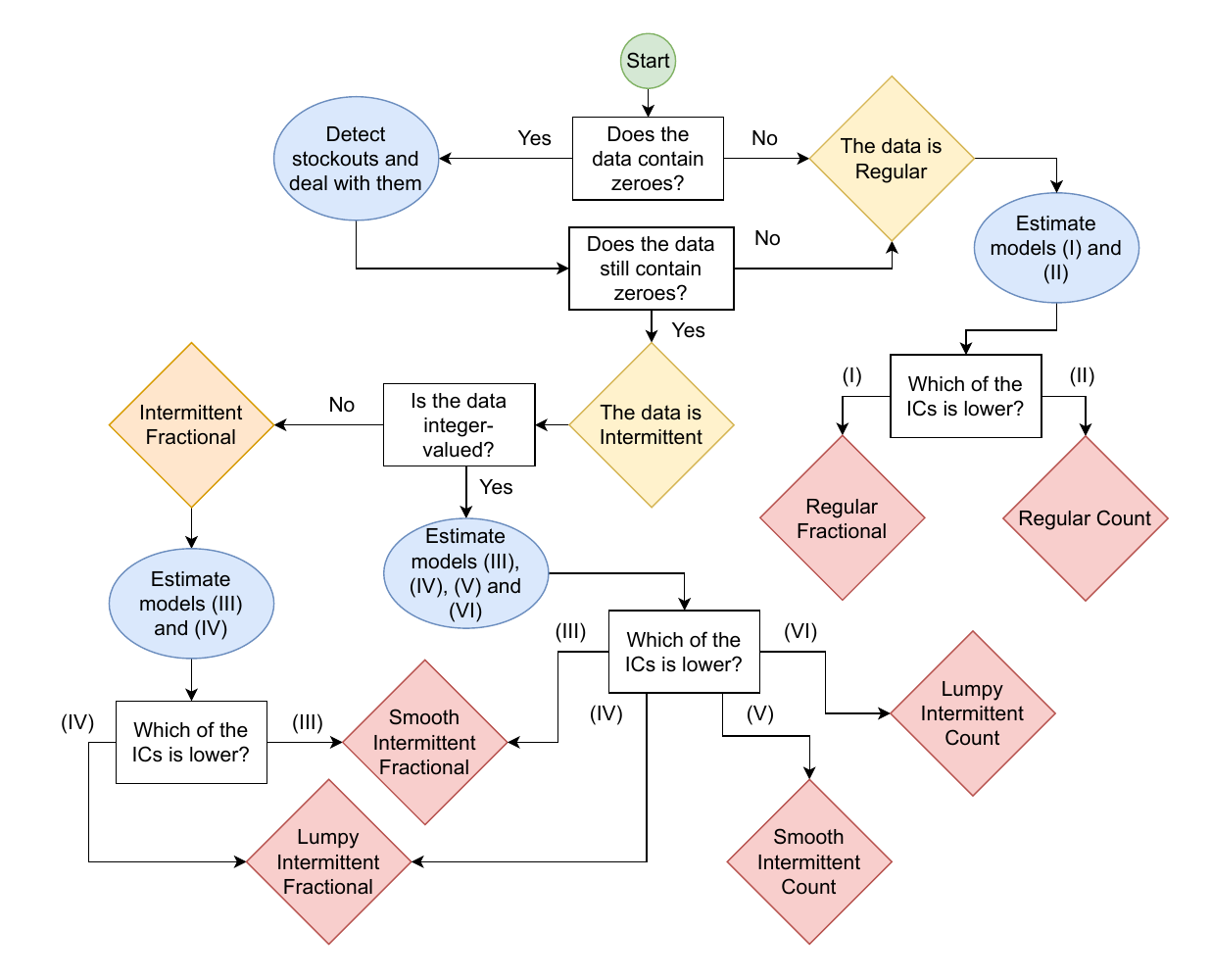}
    \caption{Flowchart of the AID algorithm. IC is the abbreviation of the term ``Information Criterion''.}
    \label{fig:AID-Algorithm-Chart}
\end{figure}

The logic in the application of the above models is as follows. In the first step, we check whether there are any zeroes left after removing stockouts to decide whether the data is regular or intermittent. If there are some left, they must be natural, meaning that the demand is indeed Intermittent. When this is decided, we then move to the next stage, calculating information criteria \citep[such as AIC by][]{Akaike1974} and comparing specific models to decide what specific type of demand we have. In case of the regular demand, we compare AIC of the model \eqref{item:RegularFrac} with the one of the model \eqref{item:RegularCount}, selecting the model, which has the lowest value. If the demand was identified as intermittent, and has fractional values, we then compare AIC of the models \eqref{item:IntermittentSmoothFrac} with \eqref{item:IntermittentLumpyFrac} to decide what specific type of demand we have: Smooth Intermittent or the Lumpy Intermittent. On the other hand, if the data has integer values only, it can be modelled using either a count data model or a fractional one. To determine, which one is better, we compare AIC of the models \eqref{item:IntermittentSmoothFrac}, \eqref{item:IntermittentLumpyFrac}, \eqref{item:IntermittentSmoothCount} and \eqref{item:IntermittentLumpyCount}. For simplicity, the whole proposed algorithm is summarised in the flowchart in Figure \ref{fig:AID-Algorithm-Chart}.

\Changes{We should also note that the AID algorithm can be extended by the inclusion of trends, seasonal components, and exogenous variables. However, we argue that the logic of the classification should not change in this case, and, for example, the model for the lumpy demand would be selected in both cases of having or not having the additional features. This is because all the models under comparison are compared conditional on the same information, and addition of features would reduce information criteria for all of them in a similar manner, thus not affecting the ranking of the models based on IC.}

%%%%%%%%%%%%%%% Simulation experiment %%%%%%%%%%%%%%%
\section{Simulation experiment} \label{sec:Simulation}
%%%%% stockouts %%%%%
\subsection{Detecting stockouts} \label{sec:Simulationstockouts}
% Perhaps, we need to simulate our data along the same lines as in \cite{Petropoulos2014}?
In the first experiment, we test the stockout detection algorithm in a scenario with both natural and artificially induced zeroes, assuming all observed stockouts are genuine. We generate data using a Geometric distribution to create demand intervals, randomly replacing some with anomalously long ones to simulate stockouts. After that we transform the intervals into the occurrence variable $o_t$, containing zeroes and ones for each observation and then substitute ones with the values from the Shifted (by one unit) Negative Binomial distribution with the probability of 0.75 and size of 5, so that all demand sizes are always positive. While it is possible to use other distributions in place of the Negative Binomial, our approach focuses on the demand intervals to detect stockouts, so it is not important what is used for the sizes.

The goal of this simulation experiment was to track how sensitive the detection mechanism is to several factors with the following expectations about their impact on performance:

\begin{enumerate}
    \item Length of stockouts -- the method should be able to detect longer stockouts easier than the shorter ones;
    \item Number of stockouts -- the method should find it harder to detect the stockouts when there are more of them in the data;
    \item Number of zeroes in the data -- its power should be inverse proportional to the overall number of zeroes in the data (inverse proportional to the probability of occurrence);
    \item Sample size -- its power should be proportional to the sample size.
\end{enumerate}

To track the performance in these dimensions, we apply several scenarios, summarised in Table \ref{tab:simulation1Setting}. While there can be many more scenarios, we wanted to make them practical, thus varying only one component and fixing the others in each one of them.

\begin{table}[htb]
    \centering
    \resizebox{1\textwidth}{!}{
    \begin{tabular}{l | c c c c}
        \toprule
        Parameters & Scenario I & Scenario II & Scenario III & Scenario IV \\
        \midrule
        Sample size & 100 & 100 & 100 & 30 -- 1000\\
        Probability of occurrence & 0.8 & 0.8 & 0.1 -- 0.9 & 0.8 \\
        Number of stockouts & 1 & 1 -- 10 & 5 & 5 \\
        Length of stockouts & 3 -- 10 & 5 & 5 & 5 \\
        \bottomrule
    \end{tabular}
    }
    \caption{The settings for the four scenarios in the first simulation experiment to track stockouts.}
    \label{tab:simulation1Setting}
\end{table}

In each of these scenarios, we track the positive and negative rates for the function by varying the confidence level, creating confusion matrices and then aggregating them for each of the setting. This way we can see how the sensitivity and specificity of our approach changes with the change of the settings inside each scenario. Using these values we create Receiver Operating Characteristic (ROC) curves, showing how well the stockouts are detected. The ideal ROC curve should be close to the left top corner, meaning that the approach always distinguishes between the true positive and true negative cases.

\begin{figure}[!htb]
	\centering
	\includegraphics[width=1\textwidth]{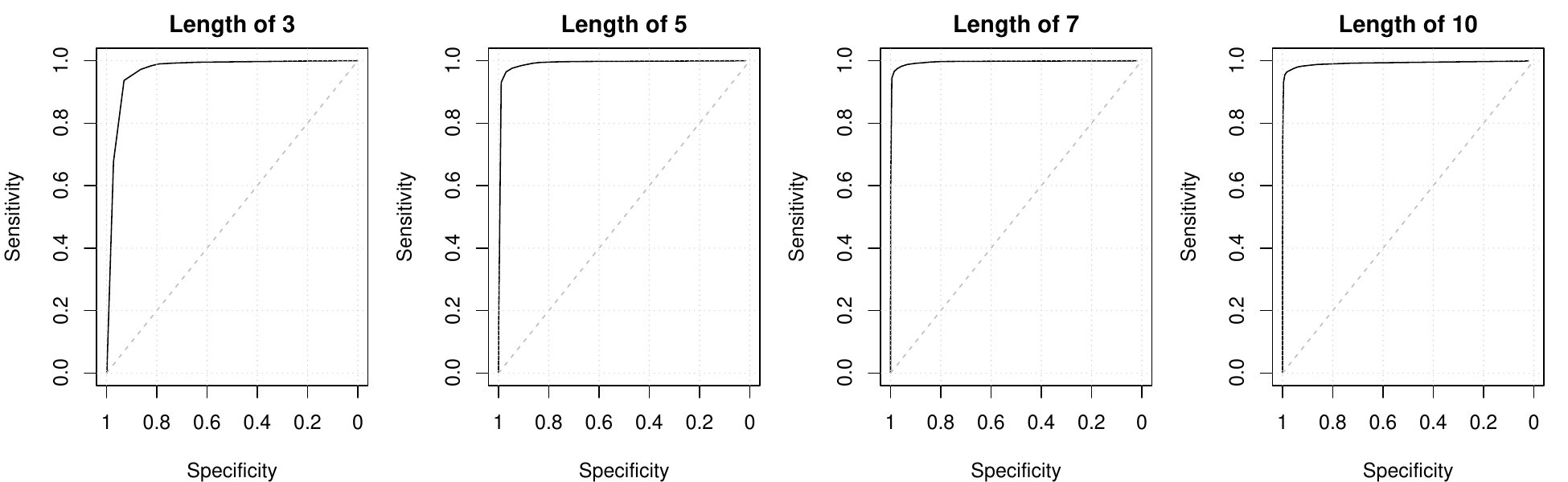}
    \caption{Scenario 1: changing the length of stockouts.}
    \label{fig:Scenario1}
\end{figure}

Figure \ref{fig:Scenario1} demonstrates the ROC curves for the Scenario 1, where the length of stockouts changes. As we see the method demonstrates lower sensitivity in case of the stockouts of length 3 in comparison with the longer ones. This is expected because in that case (when the probability of occurrence is 0.8), in the generated data, there can be slightly longer streaks of zeroes occurring naturally, and it might be hard to tell the difference between the stockout lasting for three observations and nobody buying a product for the three consecutive observations because there is no demand. With the increase of the length, it becomes easier to detect the stockouts, as we originally expected. The Area Under Curve (AUC) values for the four types of length were 0.966, 0.973, 0.976 and 0.972 respectively.

\begin{figure}[!htb]
	\centering
	\includegraphics[width=1\textwidth]{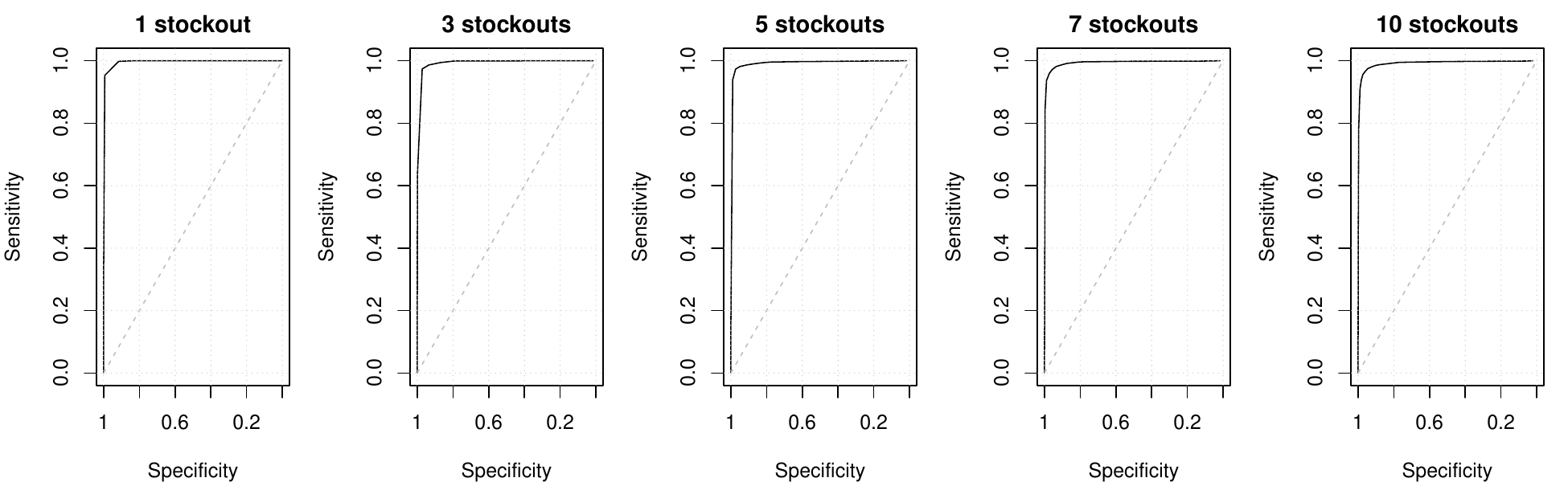}
    \caption{Scenario 2: changing the number of stockouts.}
    \label{fig:Scenario2}
\end{figure}

The results for the Scenario 2 are summarised in Figure \ref{fig:Scenario2}. While it is not very apparent from the plots, the ROC curve for 10 stockouts seems to be slightly further away from the top left corner than for the other cases. The AUC values for different stockouts number were respectively 0.996, 0.977, 0.973, 0.975 and 0.969, implying that when there are more stockouts, it become harder to detect them. This is an expected behaviour because having many stockouts makes them ``normal'' from the point of view of our approach.

\begin{figure}[!htb]
	\centering
	\includegraphics[width=1\textwidth]{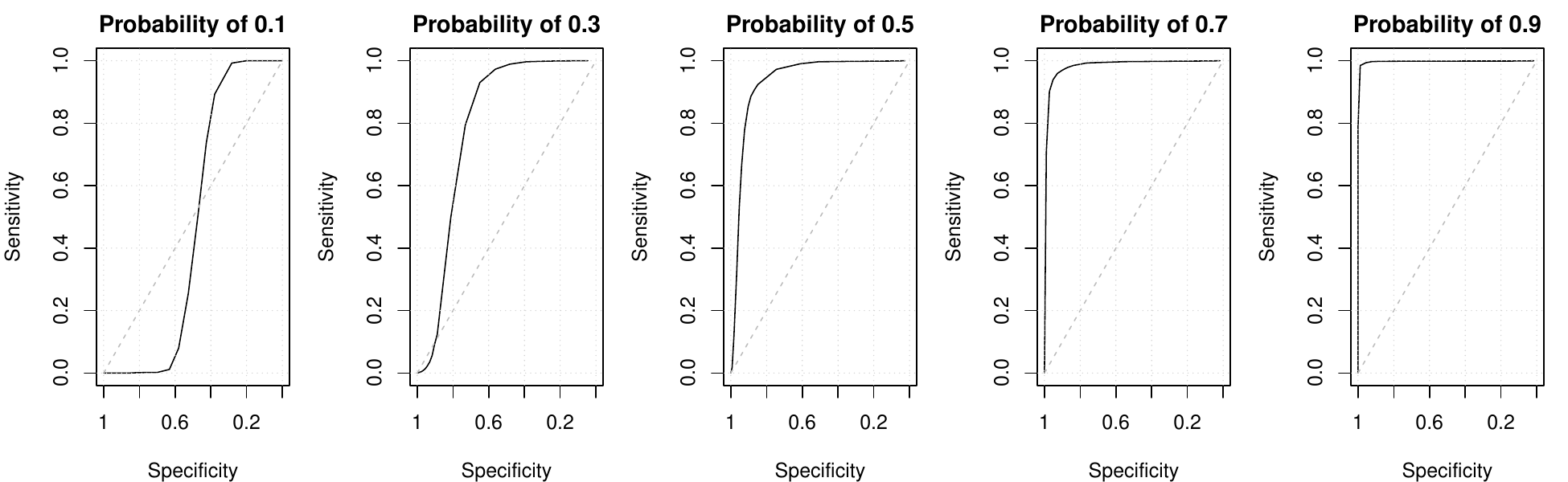}
    \caption{Scenario 3: changing the probability of occurrence.}
    \label{fig:Scenario3}
\end{figure}

The setting for the third scenario was more challenging for the approach (see Figure \ref{fig:Scenario3}): with the lower probability of occurrence it might be hard to detect stockouts because this means that there are many zeroes in the data. With the increase of probability, the approach starts working better. This is reflected in the plots in Figure \ref{fig:Scenario3}. The AUC values for this scenario were 0.473, 0.748, 0.909, 0.964 and 0.981 respectively.

\begin{figure}[!htb]
	\centering
	\includegraphics[width=1\textwidth]{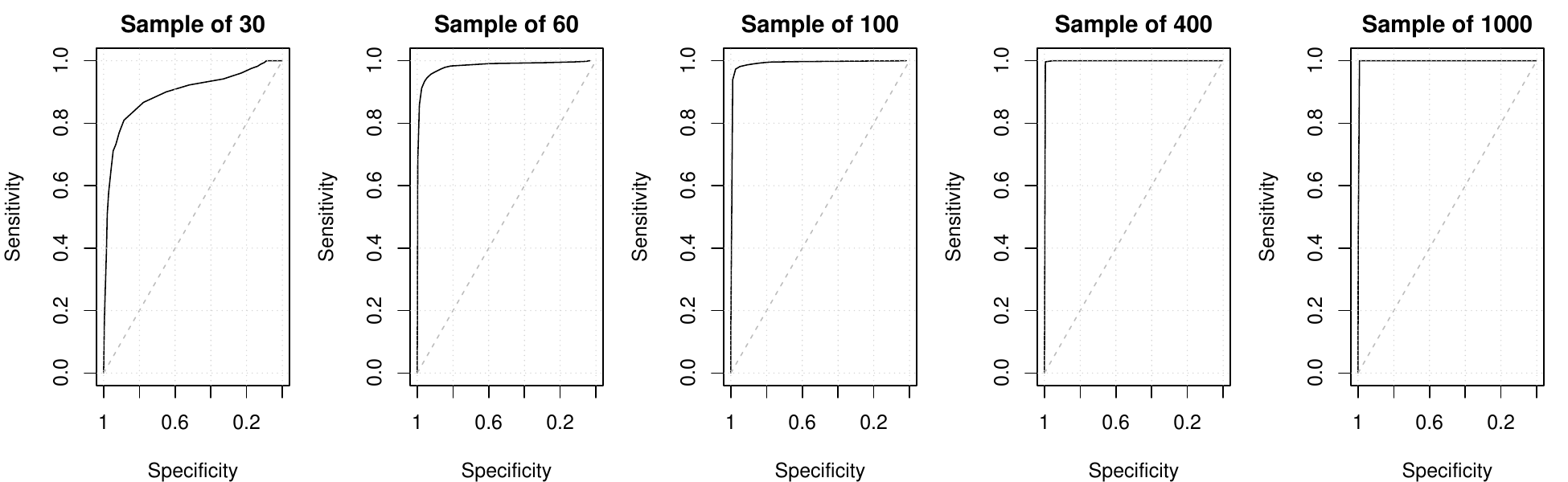}
    \caption{Scenario 4: varying sample size.}
    \label{fig:Scenario4}
\end{figure}

Finally, in Scenario 4 we varied the sample size (see Figure \ref{fig:Scenario4}). For the sample of just 30 observations, we could not have all five stockouts as planned, so we had to remove some of them. Still, detecting stockouts in such a short sample seems to be a challenging task: the ROC curve for the smallest sample is further away than for the larger ones. The best performance is achieved in the sample of 1000, which shows that the approach has enough power to detect the stockouts. The AUC values were 0.894, 0.949, 0.973, 0.994 and 0.996. While it might seem that the sample of 1000 observation is unrealistically large, some retailers keep records of data for three or more years of daily data, which can easily give more than a thousand of observations.

Summarising the results of this simulation experiment, we can see that the power of the stockouts detection approach is positively related to the sample size, probability of occurrence and the length of stockouts, and negatively related to the number of stockouts. Also we acknowledge that we simulated the easiest case of potential stockouts, so depending on the data and its granularity, our approach might produce different results, but we still argue that it could be useful as a data preparation step. 

%%%%% Demand Identification %%%%%
\subsection{Demand Identification} \label{sec:SimulationDemand}
For this part of the experiment, we simulated data from six DGPs, each representing different types of demand. These are inspired by the iETS model of \cite{Svetunkov2023b}:

\begin{enumerate}
    \item \textbf{Regular Fractional}: ETS(M,N,N) with the Log-Normal distribution of the residuals. We used the multiplicative error model to make sure that the generated data is positive. The initial level was set to 1000;
    \item \textbf{Smooth Intermittent Fractional}: ETS(A,N,N) with the normal distribution and the initial level of 10. After generating the data, all negative values were substituted by zeroes. This aligns with the Rectified Normal distribution discussed above;
    \item \textbf{Lumpy Intermittent Fractional}: ETS(M,N,N), similar to (1), but after generating the data, random zeroes were introduce (so that 30\% of observations are zero);
    \item \textbf{Regular Count}: First, the data was generated using ETS(M,N,N) with the same parameters as in (1), after which it was used in the data generation from the Negative Binomial distribution with size 20 and the mean equal to the ETS(M,N,N) data. This way, the level of series would evolve over time, but the values themselves will be count;
    \item \textbf{Smooth Intermittent Count}: Similar to (4), but with lower initial level (5 instead of 10) and lower size (2 instead of 20). This way the data will also have some occasional zeroes;
    \item \textbf{Lumpy Intermittent Count}: Similar to (4), but introducing random zeroes (30\% of them).
\end{enumerate}

The data was generated using the \texttt{sim.es()} function from the \texttt{smooth} package in R \citep{SvetunkovSmooth}. The resulting series looked similar to the data shown in Figure \ref{fig:Categories-Full}. The simulation was done for the samples of 30, 60, 100, 400 and 1000 observations.

We then applied the \texttt{aid()} function with a confidence level of 0.999 for stockout detection and recorded how often each demand category was correctly identified. Using such a high confidence level reduces the likelihood of zeroes being misclassified as stockouts, though the AID approach may still occasionally flag some incorrectly. Lowering the confidence level would result in more intermittent series being classified as regular, as more zeroes would be treated as artificially occurring. Another option would be to set the level equal to one, thus switching off the stockouts detection mechanism completely, but we decided not to do that because we wanted to see the impact of the mechanism on the final classification.

\begin{figure}[!htb]
	\centering
	\includegraphics[width=\textwidth]{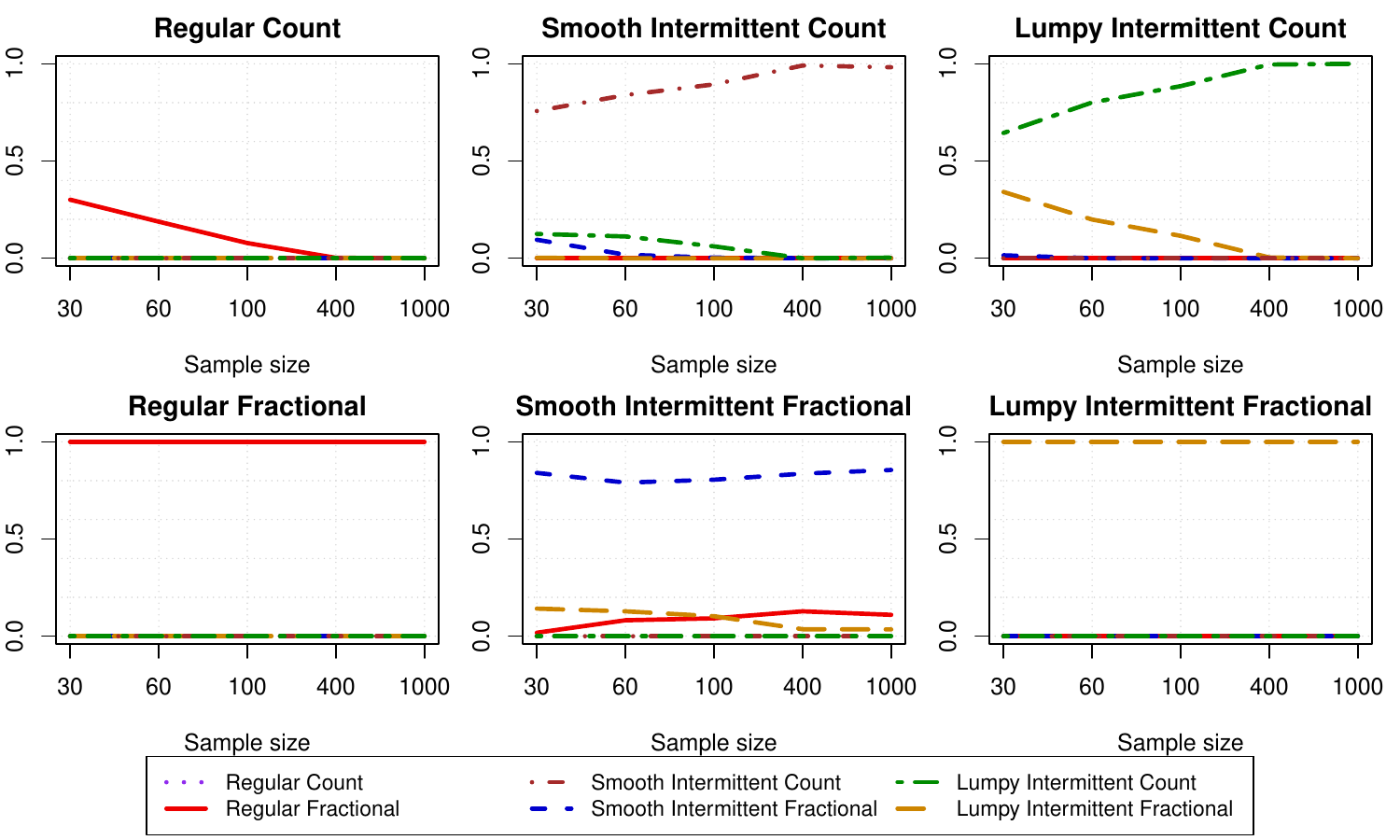}
    \caption{Demand identification for each of categories of DGPs.}
    \label{fig:IDGeneral}
\end{figure}

The results of this simulation experiment are shown in Figure \ref{fig:IDGeneral} with lines representing the percentage of demands identified for each of DGPs. We can see that the identification of ``Lumpy Intermittent Fractional'' and ``Regular Fractional'' is done with 100\% precision for any sample size. This is because these types of demand are very special and easy to identify. With all the other categories, the algorithm struggled on small samples and then became more powerful, being able to identify demand correctly on larger samples.

Overall, the algorithm struggled to correctly identify count data, particularly with smaller samples. This is because fractional demand models can often be applied effectively, even when the data is count-based \citep[as shown, for example, by][]{Svetunkov2023b}. The algorithm also had difficulty identifying ``Smooth Intermittent Fractional" demand, even with the largest sample of 1000 observations. This was mainly due to the AID approach frequently flagging zeroes as stockouts for this data, causing the demand to be misclassified as ``Regular Fractional". In some cases, series in this group were also misclassified as ``Lumpy Intermittent Fractional", although this misclassification decreased with the increase of the sample size. This is likely because models for lumpy intermittent demand can also be effectively applied to smooth intermittent series.

\begin{figure}[!htb]
	\centering
	\includegraphics[width=\textwidth]{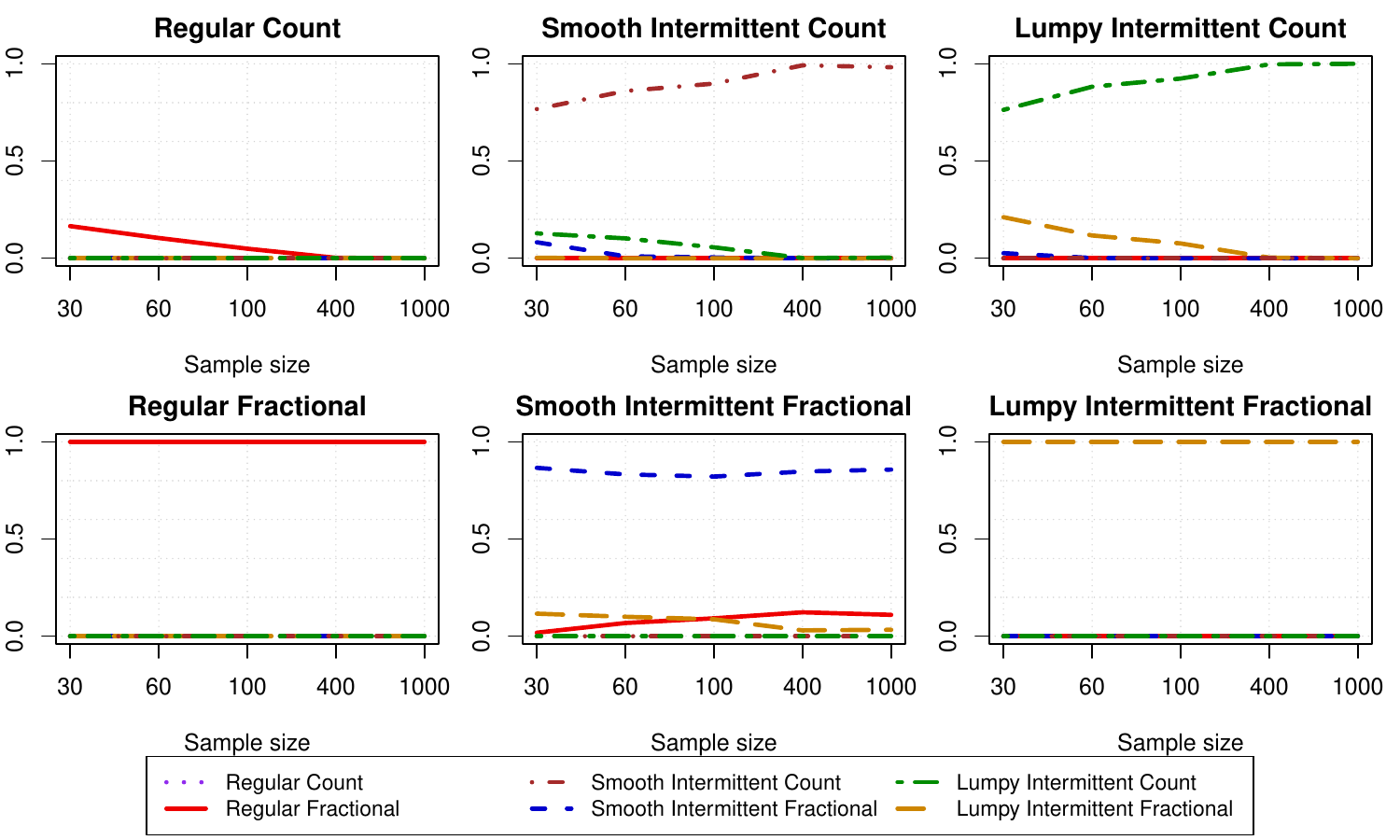}
    \caption{Demand identification for each of categories of DGPs. The case of missed promotions.}
    \label{fig:IDGeneralPromo}
\end{figure}

\Changes{In addition to this, we also conducted an experiment where outliers are added to the data, reflecting the situation of promotions that are not taken into account by the AID algorithm. They were added randomly in the sample when non-zero values occurred by multiplying those values by two. The idea was to see how the results would change if we have some elements of the structure that are ignored by AID. The results of this additional experiment are shown in Figure \ref{fig:IDGeneralPromo}.}

\Changes{Comparing Figures \ref{fig:IDGeneral} and \ref{fig:IDGeneralPromo}, we can see that the only thing that changes in this situation is the accuracy of the identification of the fractional vs count demand on small samples: the demand in those two categories is now identified better than in the case of no promotions. This is probably because the multiplication by two leads to values that look more apparent as fractional/count.}

In summary, the simulation experiment shows that the proposed approach performs well with larger samples but makes some mistakes with smaller ones. It effectively detects regular and lumpy intermittent demand, though it struggles with the smooth category due to limitations in the stockout detection algorithm. The distinction between count and fractional demand is not particularly pronounced from a modelling perspective, which leads the AID method to occasionally misclassifying the count ones. Nevertheless, the approach serves as a solid starting point for further analysis. To improve its accuracy, the confidence level for stockout detection should be carefully adjusted to prevent naturally occurring zeroes from being flagged as stockouts. Additionally, if identifying count data is particularly important, it is advisable not to rely solely on automated detection but to directly verify whether the demand sizes are integer-valued.

%%%%%%%%%%%%%%% Example %%%%%%%%%%%%%%%
\section{Case study} \label{sec:CaseStudy}
\subsection{Experiment setting}
To assess the proposed classification scheme, we used sales data of a retailer. This contained 342 weekly observations, starting from 1st April 2018 and finishing on 4th November 2024 with some products having shorter histories than the others. The dataset contained 3 shops with overall 31018 products. The task at hand was to produce forecasts for two weeks ahead, so we withheld the last two observations to check the accuracy of applied approaches. The main idea was to understand whether the proposed stockouts detection and then demand classification algorithm would improve the accuracy of forecasting approaches. We note that the forecasting methods used in this paper were selected by the authors, although the company uses similar approaches. We cannot disclose any specific details due to the Non-Disclosure Agreement.

%%%%% stockouts detection %%%%%
\subsection{Stockouts detection} \label{sec:CS_stockouts}
\begin{figure}[!htb]
    \centering
    \includegraphics[width=0.85\textwidth]{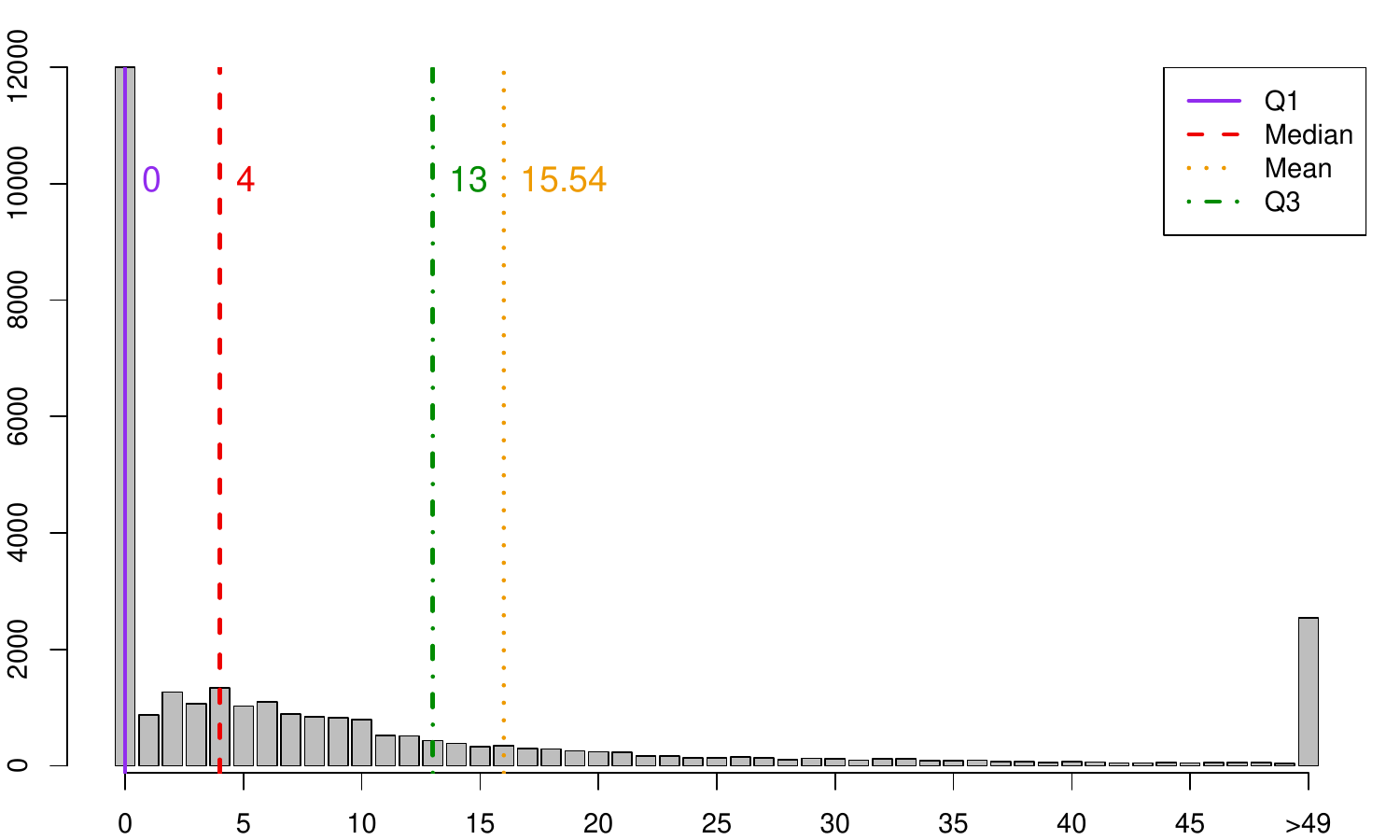}
    \caption{Distribution of number of stockouts per series for the retail data.}
    \label{fig:AID-Stockouts}
\end{figure}

We applied our classification scheme to the data via the \texttt{aidCat()} function from the \texttt{greybox} package in R \citep{Svetunkovgreybox} with \texttt{level=0.999}. It identified stockouts for each time series, the distribution of which is summarised in Figure \ref{fig:AID-Stockouts}.

As we see, around 12,000 time series did not have any stockouts. But when they happened for the other series, in the majority of cases, there were from one to 10 gaps in the data. The half of time series had fewer than four stockouts, but there were also some that had many more zeroes. The visual inspection of some extreme cases revealed, that some products were sold in 2018 but then were discontinued until the end of 2024.

We do not have any additional information from the company, so we cannot conclude whether the detection mechanism worked well, but we will use the detected stockouts as features in the models in the next Subsections to see whether they bring improvements in terms of accuracy.

%%%%% AID groups %%%%%
\subsection{Demand categories}
After applying the AID algorithm, we ended up with the 6 demand categories shown in Table \ref{tab:Rewe-ID-categories}. We checked the algorithm with other significance levels (0.99 and 0.9999), but found that the results do not change substantially.

\begin{table}[!htb]
    \centering
    \resizebox{1\textwidth}{!}{
    \begin{tabular}{r | c c c | c}
        \toprule
        & Regular & Smooth Intermittent & Lumpy Intermittent & Overall\\ 
        \midrule
        Count & 5652 & 19128 & 2753 & 27533\\ 
        Fractional & 1115 & 1342 & 1016 & 3473 \\ 
        \midrule
        Overall & 6767 & 20470 & 3769 & 31006 \\
        \bottomrule
    \end{tabular}
    }
    \caption{Demand classification for the retail company data.}
    \label{tab:Rewe-ID-categories}
\end{table}

We see that the majority of time series were flagged as ``Smooth Intermittent Count'', around 20\% of them were ``Regular'' and only 10\% were flagged as ``Lumpy Intermittent'' This means that we are dealing with the fairly homogenous dataset, and the effect of the features for each category might be not very well pronounced.

\begin{figure}[htb]
    \centering
    \includegraphics[width=1\textwidth]{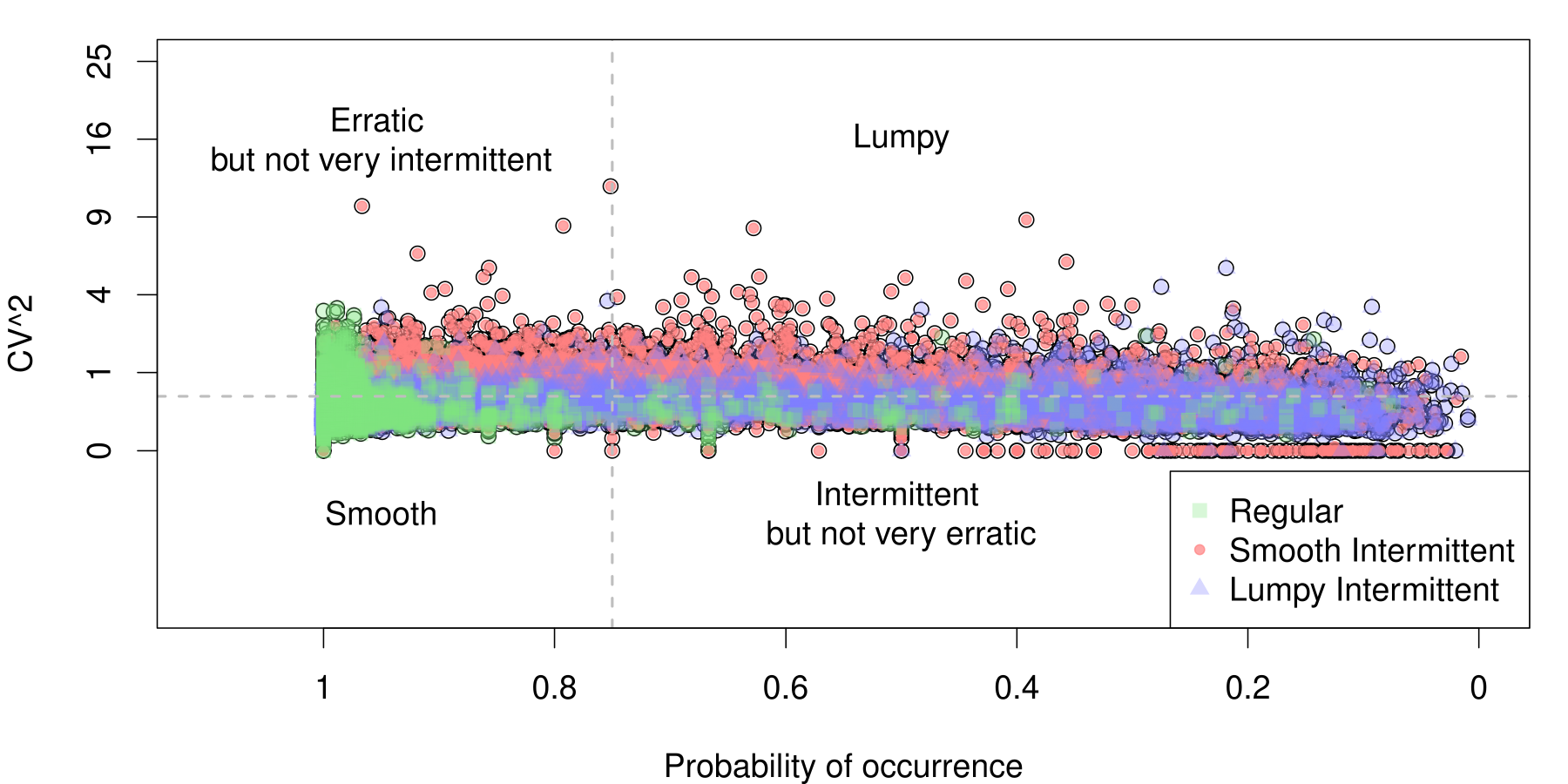}
    \caption{Demand classification for the retail company data according to SBC and AID classifications. The SBC is depicted in four quadrants, while the AID depicts dots in green (squares), red (circles) and blue (triangles) for the Regular, Intermittent Smooth and Intermittent Lumpy demands respectively.}
    \label{fig:Rewe-SBC-AID}
\end{figure}

Furthermore, we decided to compare classifications according to SBC and AID to better understand whether they have anything in common. The visualisation of the two approaches is shown in Figure \ref{fig:Rewe-SBC-AID}. We see that AID produces a classification based on a non-linear split, while SBC just separates the space into four quadrants. The thing to note is that some of time series flagged as ``Regular'' (green squares) according to our classification were categorised as Intermittent according to SBC. This is because SBC does not treat stockouts and if those zeroes are removed, the data would indeed become regular.

Furthermore, we argue that while the SBC seems more convenient, it is less useful for model selection and feature engineering if it is done for the modern approaches. This is because, as we discussed earlier, it was originally developed for the selection between Croston's and SBA methods. The AID approach, on the other hand, seems less straightforward, but it relies on the suitability of models rather than arbitrary data characteristics. So, arguably, AID can be used outside of the classical intermittent demand forecasting methods and can be applied to a wider range of more modern techniques.

%%%%% Forecasting approaches and features %%%%%
\subsection{Forecasting approaches and features}
Given that we were not interested in finding the most accurate forecasting approach, we wanted to see the effect of the AID algorithm on forecasting, we used several approaches:
\begin{enumerate}[I.]
    \item \textbf{LightGBM} \citep{Ke2017} -- because of its speed and ability to handle large datasets like the one we had. We did our experiments using the \texttt{lightgbm} package in R \citep{ShiLightGBM};
    \item \textbf{Pooled regression} applied to the whole dataset -- to see whether the findings hold for a linear model that does not have as much flexibility as the LightGBM. This was done using the \texttt{alm} function from the \texttt{greybox} package in R \citep{Svetunkovgreybox};
    \item \textbf{Smoothed} series -- a variety of smoothed series, aiming at capturing the local level using several options discussed later in this subsection. With these, we wanted to see whether there was an added benefit in treating different time series features locally, per series.
\end{enumerate}

For the decision tree approach, we used several features collected by the company:

\begin{enumerate}
    \item \label{list:promo} Promotions -- dummy variables, indicating when an item was on promotion;
    \item \label{list:holidays} Holidays -- categorical variables, denoting holidays, such as:
    \begin{multicols}{2}
        \begin{itemize}
            \item All Saints;
            \item Ascension of Christ;
            \item Corpus Christi;
            \item Holy Three Kings;
            \item Easter Sunday;
            \item Maria Conception;
            \item Assumption Day;
            \item National Holiday;
            \item New Year;
            \item Easter Monday;
            \item Pentecost;
            \item Labour Day;
            \item Christmas Day;
            \item St. Stephen's Day;
        \end{itemize}
    \end{multicols}
    \item \label{list:events} Events -- categorical variables, denoting special events for specific dates, such as:
    \begin{multicols}{2}
        \begin{itemize}
            \item Shrove Tuesday;
            \item Mother's Day;
            \item St. Nicholas Day;
            \item St. Valentine's Day;
            \item Father's Days;
        \end{itemize}
    \end{multicols}
    \item \label{list:covid} Covid -- a binary variable capturing the effect of covid on sales;
\end{enumerate}

We also devised our own features that should improve the accuracy of the forecasting approaches:
\begin{enumerate}
	\setcounter{enumi}{4}
    \item \label{list:stockout} Stockout -- a dummy variable, showing when stockouts happened according to our approach with the confidence level of 0.999. These were discussed in Subsection \ref{sec:CS_stockouts};
    \item \label{list:category} \Changes{Category Partial -- the demand categories according to AID, including only either ``intermittent'', or ``regular''. The idea of this feature is to see whether the split to the these two categories would improve the performance of models};
    \item \label{list:categoryFull} \Changes{Category Full -- the three demand categories according to AID, including ``regular'', ``smooth intermittent'' and ``lumpy intermittent''. Similarly to the previous feature, this one explores whether the further split brings value};
    \item \label{list:smoothSales} SmoothSales -- Smoothed original series. The smoothing was done using Friedman's Super Smoother \citep{friedman1984variable} via the \texttt{supsmu()} function from the \texttt{stats} package in R \citep{RCore2020}. We used it because it is more sensitive to the local level changes than LOWESS. If the smoothed line was below zero, we would substitute its values with zero, reflecting the fact that the sales cannot be negative. In cases of small samples (less than 7 non-zero observations), we substituted the values by the in-sample mean. For the holdout part of the data, we repeated the last available smoothed value in the sample for each series;
    \item \label{list:smoothDemand} SmoothDemand -- Another version of the smoothed series, done by excluding the observations that were detected as stockouts using our approach. The resulting gaps in the smoothed series were interpolated linearly. This way we would capture the true level of demand, instead of sales. The forecasts from this are done similarly to \eqref{list:smoothSales};
    \item \label{list:smoothSizes} SmoothDemandSizes -- Furthermore, we smoothed the demand sizes only (dropping all the zeroes), which was an important feature for the mixture model (see below);
    \item \label{list:probability} Probability -- the smoothed binary demand occurrence variable (the estimate of the probability of occurrence), done after removing the stockouts. If the smoothed line was above one or below zero, we would substitute its values with the boundary values;
    \item \label{list:seasonality} Seasonality -- sine and cosine curves with periodicity of 52 (i.e. $\sin\left(\frac{2\pi j}{52}\right)$) were added as well. These help capturing the week-of-year seasonality by non-linear machine learning approaches (such as LightGBM), but are not useful for linear ones, such as regression \citep{Barrow2015}.
\end{enumerate}

\Changes{We remark that we do not add the feature of count/fractional demand because our experiments showed that while it is theoretically appealing, it does not bring any value in terms of forecasting accuracy.} Furthermore, while there can be many other features that could be added to the experiment (such as ETS components or smoothed quantiles of the data), the aim was not to find the most suitable set of features, but rather to better understand the impact of our classification on forecast accuracy.

To see improvements brought by the introduction of our features, we evaluated five approaches:

\begin{enumerate}[A.]
    \item \label{list:Conventional} \textbf{Conventional} -- one LightGBM approach applied directly to the full dataset ignoring the stockouts feature and using features \eqref{list:promo}, \eqref{list:holidays}, \eqref{list:events}, \eqref{list:covid}, \eqref{list:smoothSales} and \eqref{list:seasonality};
    \item \label{list:Full} \textbf{Full} -- similar to \eqref{list:Conventional}, but with the stockouts dummy variable \eqref{list:stockout} and feature \eqref{list:smoothDemand} instead of \eqref{list:smoothSales};
    \item \label{list:Mixture} \textbf{Mixture} -- two approaches applied to the dataset after splitting every observation into demand occurrence and demand sizes via equation \eqref{eq:CrostonIdea}. The LightGBM applied to the former focused on predicting the probability of occurrence, while the latter one focused on predicting the demand sizes. The former used features \eqref{list:promo}, \eqref{list:holidays}, \eqref{list:events}, \eqref{list:covid}, \eqref{list:stockout}, \eqref{list:seasonality}, and \eqref{list:probability}, while the latter had \eqref{list:promo}, \eqref{list:holidays}, \eqref{list:events}, \eqref{list:covid}, and \eqref{list:smoothSizes}. After that, the forecasts from the two were combined via the multiplication to get the final values;
    \item \label{list:CategoryPartialModel} \Changes{\textbf{Category Partial} -- Same as the Mixture approach, but with the feature \ref{list:category} added to both demand occurrence and the demand sizes parts of the approach;}
    \item \label{list:CategoryFullModel} \Changes{\textbf{Category Full} -- Same as the Category Partial approach, but with the feature \ref{list:categoryFull} instead of \ref{list:category}.}
    % \item \label{list:CategoryPartial} \textbf{Category Partial} -- three LightGBMs, one applied to the data which was flagged as ``Regular'' in the manner similar to \eqref{list:Full}, and the other two applied to the data flagged as ``Intermittent'' in the manner similar to \eqref{list:Mixture};
    % \item \label{list:CategoryFull} \textbf{Category Full} -- Similar to \eqref{list:CategoryPartial}, but with the split of the data into Regular/Smooth Intermittent/Lumpy Intermittent. The ``Full'' approach was applied to the Regular demand and two separate ``Mixture'' approaches were used for the Smooth and Lumpy intermittent demand. With this split, we want to see whether the more thorough split into categories brings any improvement;
    % \item Local Level -- just a smoothed series with a straight line for the holdout, similar in style with the forecast from the moving average, aiming to capture the local level of each series. This was used as a benchmark to understand whether LightGBM brings value at all.
\end{enumerate}

The logic in fitting the approaches above was to investigate the following three aspects:

\begin{itemize}
    \item The effect of stockout detection mechanism on the accuracy by comparing performance of approaches \eqref{list:Conventional} and \eqref{list:Full};
    \item The impact on the accuracy of the mixture approach that splits the data into the demand occurrence and demand sizes parts by comparing approaches \eqref{list:Full} and \eqref{list:Mixture};
    \item The usefulness of the proposed demand classification by comparing \eqref{list:CategoryPartialModel} and \eqref{list:CategoryFullModel} with \eqref{list:Full} and \eqref{list:Mixture}.
\end{itemize}

% We should note at this stage that we could not come up with unique features that would support the split into smooth and lumpy intermittent categories. The separation into two has a reasonable theoretical rationale but does not yet has distinct characteristics.

Furthermore, we applied pooled regression in the manner similar to the LightGBM to make sure that the main idea of the paper holds irrespective of the used approach. We use the same naming convention as in case of the LightGBM. In the process of model fitting we noticed that any company feature degrades the accuracy of pooled regression, so we dropped them. The only features that bring value were the ones that we generated, including the stockouts dummy variable.

We also show the accuracy of the smoothed series \eqref{list:smoothSales}, \eqref{list:smoothDemand}, and \eqref{list:smoothSizes} combined with \eqref{list:probability}, keeping the same names as for the LightGBM and regression. We note that in case of the regular demand, the probability of occurrence was equal to one, and the smoothed line \eqref{list:smoothDemand} should coincide with \eqref{list:smoothSizes}.

We should note that in this section, we only focused on measuring the point forecast accuracy of approaches, by calculating the Root Mean Squared Scaled Error (RMSSE) from \cite{Makridakis2022}, originally motivated by \cite{Athanasopoulos2020}. But an important aspect in the evaluation is that \textbf{it needs to be done on the test set values that do not have stockouts}. If the stockouts are not removed from the test set, the evaluation would point towards the models that are the most accurate at predicting sales instead of demand. We want to avoid that, so we dropped series that had only zeroes in the test set.

\Changes{Finally, we conducted additional experiment using the count/fractional as a feature in the models, but the performance of LightGBM and regression was similar to the ``Category Partial'' of the respective models. We argue that the fact that the demand is either fractional or count does not matter that much for the forecasting accuracy, but it might be important for inventory decisions if we deal with a mixture of products.}

%%%%% Results %%%%%
\subsection{Forecasting results} \label{sec:resultsPoint}

The results of this experiment are summarised in Table \ref{tab:LightGBM}. There are several takeaways from it:

\begin{table}[ht]
    \centering
    \resizebox{1\textwidth}{!}{
    \begin{tabular}{r | cccccc}
        \toprule
        & min & Q1 & median & mean & Q3 & max \\
        \midrule
        \multicolumn{7}{c}{LightGBM} \\
        \midrule
        Conventional    & 0.0026 & 0.4849 & 0.7966 & 1.1242 & 1.3113 & 72.1749 \\ 
        Full            & 0.0056 & 0.4754 & 0.7780 & 1.0907 & 1.2788 & 70.4951 \\ 
        Mixture         & 0.0000 & 0.4550 & 0.7574 & 1.0682 & 1.2375 & 81.7721 \\ 
        Category Partial & 0.0000 & \textbf{0.4288} & 0.7125 & 1.0008 & 1.1624 & 62.9427 \\ 
        Category Full   & 0.0000 & 0.4296 & \textit{\textbf{0.7107}} & \textit{\textbf{0.9982}} & \textit{\textbf{1.1600}} & \textit{\textbf{62.8027}} \\ 
        \midrule
        \multicolumn{7}{c}{Pooled Regression} \\
        \midrule
        Conventional    & 0.0007 & 0.5057 & 0.8249 & 1.1669 & 1.3433 & 105.3349 \\ 
        Full            & 0.0004 & 0.5158 & 0.8244 & 1.1492 & 1.2852 & 251.5448 \\ 
        Mixture         & 0.0000 & 0.4597 & 0.7657 & 1.0766 & 1.2486 & 82.0593 \\ 
        Category Partial & 0.0000 & \textit{\textbf{0.4255}} & \textbf{0.7125} & 1.0016 & 1.1623 & \textbf{78.1823} \\ 
        Category Full   & 0.0000 & 0.4275 & 0.7142 & \textbf{1.0006} & \textbf{1.1622} & 79.5183 \\ 
        \midrule
        \multicolumn{7}{c}{Smoothed Series} \\
        \midrule
        Conventional    & 0.0000 & 0.5023 & 0.8148 & 1.1481 & 1.3198 & 82.2092 \\ 
        Full            & 0.0000 & 0.5042 & 0.8148 & 1.1427 & 1.3091 & 82.2092 \\
        Mixture         & 0.0000 & \textbf{0.4565} & \textbf{0.7600} & \textbf{1.0686} & \textbf{1.2382} & \textbf{82.0491} \\ 
        % Category Partial & 0.0000 & 0.4638 & 0.7728 & 1.0834 & 1.2596 & \textbf{82.0491} \\
        \bottomrule
    \end{tabular}
    }
    \caption{RMSSE values of forecasting approaches with different features on the retail company data. Q1 and Q3 are the first and third quartiles respectively. The lowest error measures in each category are shown in boldface, the lowest overall are in boldface and italic.}
    \label{tab:LightGBM}
\end{table}

\begin{itemize}
    \item All LightGBM methods are more accurate than the conventional smoothed series (those that ignore stockouts) across all statistics of the RMSSE;
    \item The approach that has the smoothed series without stockouts and a separate stockouts feature (entitled ``Full'' in the table) performs better than the Conventional one applied to the dataset without the stockouts feature. This applies for LightGBM, Pooled Regression, and the Smoothed Series, showing that it is the principle of capturing the demand instead of the sales, which plays the crucial role in accuracy improvements;
    \item The split into demand occurrence and demand sizes (``Mixture'') leads to further improvements in terms of RMSSE in comparison with the ``Full'' model for all approaches with a substantial decreases in RMSSE;
    \item \Changes{The introduction of the Category feature (which only has ``regular/intermittent'' ones) further improves the accuracy both in case of the LightGBM and regression;}
    \item \Changes{The split into regular/smooth intermittent/lumpy intermittent leads to further improvement in accuracy (``Category Full''), except for the median RMSSE in case of regression.}
    % \item The split into Regular/Intermittent categories (``Category Partial'') does not bring improvements in comparison with the ``Mixture'' approach. In all three cases, such split increases the mean and quartile RMSSE.
    % \begin{itemize}
    %     \item In case of the LightGBM, the mean and median RMSSE go up. The maximum value of the RMSSE decreases, which implies that the approach does not do as big mistakes as the previous one. This is useful in practice where very poor performance of approaches on some observations can raise serious concerns of the data scientist team;
    %     \item In case of regression, there is a further improvement in terms of mean and quantiles of RMSSE, which could be attributed to having a more flexible split into categories of data;
    %     \item As for the smoothed series, there is no apparent improvement, and it seems that the split harms the accuracy;
    % \end{itemize}
    % \item The split into finer categories of Regular/Smooth/Lumpy (``Category Full'') harms LightGBM, at the same time slightly improving performance of Pooled Regression in comparison with the ``Category Partial''.
    % does not lead to any consistent noticeable improvements in comparison with the simpler classification (``Category Partial'') for LightGBM, but the Pooled Regression seems to produce more accurate forecasts in terms of mean, median, Q1 and Q3 in that case.
\end{itemize}

% We also analysed what values the occurrence part of the Mixture model produced as forecasts, and noticed that LightGBM produced all values close to the bounds (i.e. either very close to zero or to one). This was due to the "SmoothDemandSizes" feature, dropping which would result in a continuum of probabilities, not a bimodal distribution of values. So, the occurrence part of the model in this situation starts acting as a hard filter, making the final forecast either equal to zero or to the demand sizes.

\begin{figure}[htb!]
    \centering
    \includegraphics[width=\textwidth]{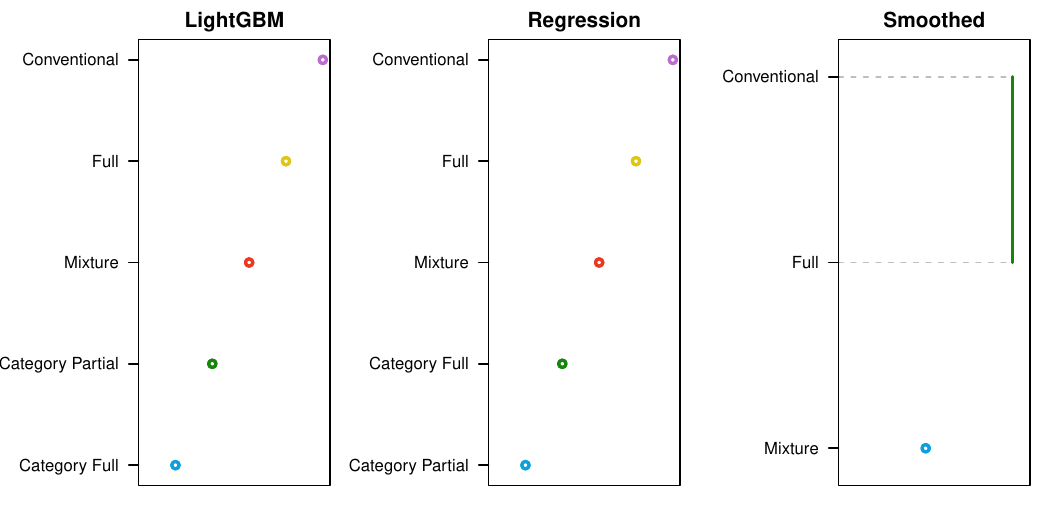}
    \caption{Nemenyi test for the LightGBM and Regression approaches. The vertical lines indicate approaches that are not statistically different on the 5\% level.}
    \label{fig:Rewe-Nemenyi}
\end{figure}

We conducted the Nemenyi test \citep{Demsar2006} implemented in the \texttt{rmcb()} function in the \texttt{greybox} package in R \citep{Svetunkovgreybox} to see whether the differences in the performance of the approaches is statistically significant on the 5\% level. Figure \ref{fig:Rewe-Nemenyi} depicts the results of this test, showing the average ranks for each of the approaches on the y-axis: the lower the approach is located, the higher rank it has, meaning that it outperforms the others on series-to-series basis more often than the other ones. If the differences in performance between approaches is not significant on the 5\% level, the vertical line is drawn, connecting them. If the differences are significant, the dot is placed in the plot. Figure \ref{fig:Rewe-Nemenyi} shows performance of approaches per group.

\Changes{For the LightGBM group, we see that the ``Category Full'' outperforms the others in the majority of cases, resulting in the highest rank. In fact, we see that the ranking observed in Table \ref{tab:LightGBM} is preserved, and the main observations discussed above stay.}

\Changes{For the Pooled Regression, the ``Category Partial'' dominates the others in terms of accuracy. Although in terms of the mean RMSSE, ``Category Full'' was slightly better than the ``Category Partial'', on the individual level of series, it does better.}
% We can also see that the main idea holds: the stockouts inclusion improves performance, split into demand occurrence and sizes leads to further improvement, and finally the introduction of the categories as features leads to the best performance of the regression model.

\Changes{For the Smoothed series, the ``Mixture'' was ranked the highest on average, outperforming the others significantly on the 5\% level. Note that the ``Conventional'' is not statistically different from the ``Full'' category in this case. This shows that the main benefit comes not from the smoothed series, but rather from the stockout feature itself, which cannot be included in case of just Smoothed time series.} 

Most importantly, the ordering of ``Category'', ``Mixture'', ``Full'', ``Conventional'' is roughly preserved for the three approaches, implying that the respective features indeed bring value.

Summarising the results of this experiment, we see that there is a value in detecting stockouts and including them in a forecasting approach (as long as they are removed from the smoothed series) and that the split into the demand sizes and demand occurrence tends to substantially improve performance of forecasting approaches. \Changes{The split into the Regular/Intermittent categories, which is done automatically after detecting stockouts, tends to further improve performance. Finally, the split into finer categories of demand further reduces significantly forecast errors in case of LightGBM and helps substantially in the pooled regression.} The latter can be attributed to the fact that regression is a linear model, and a split into finer categories makes it more flexible.

\subsection{\Changes{Inventory simulation}} \label{sec:InventorySimulation}
\Changes{To better understand whether the usage of the AID approach translates to better inventory decisions, we setup a simple inventory simulation for the available data. We did not have the information about the prices of products, so we simplified the whole setup to the following:
\begin{itemize}
    \item we used the order-up-to-level policy,
    \item the lead time was set to one week, because for the majority of products it was on average close to one,
    \item we assume that all the products expire after one week. While this is a strong assumption for so many SKUs, we did not have the information about the expiration dates, and we argue that this assumption does not violate the main findings of the experiment,
    \item we do the simulation with rolling origin for two steps, not updating the features (i.e. not recalculating the smoothed series),
    \item we used service levels of 90\%, 95\% and 99\% recording the achieved service level, scaled lost sales and inventory on hand similarly to how it was done by \cite{Kourentzes2019}.
\end{itemize}}

\Changes{To get the safety stock levels, we used a rather simple approach, relying on the empirical quantile calculation, similar to how it was done by \cite{Lee2014} and then adapted to inventory management by \cite{Trapero2019} (nowadays approaches like this are sometimes called ``conformal prediction''). In nutshell, the approach can be summarised in following steps:
\begin{enumerate}
    \item Take a LightGBM approach used in the previous section;
    \item Apply it to the training data to get the fitted values $\hat{y}_t$;
    \item Generate forecast errors $e_t = y_t - \hat{y}_t$;
    \item Scale forecast errors by the standard deviation of each separate time series;
    \item Extract the necessary quantiles of the forecast errors;
    \item Scale back produced quantiles for each series by multiplying them by standard deviations;
    \item Produce point forecasts for each time series (this was done in Subsection \ref{sec:resultsPoint});
    \item Add quantiles to each of the point forecasts to get the safety stock levels;
    \item Round up the final values, because we deal with integer-valued data.
\end{enumerate}}

\Changes{The advantage of this simple approach is that it is fast and does not require retraining any of the approaches. The disadvantage is that it relies on the in-sample forecast errors, which might lead to potential overfitting and thus lead to lower quantiles than needed. But we argue that it should suffice for the demonstration purposes: if we see the same sequence in improvement for the approaches as we saw in Subsection \ref{sec:resultsPoint}, this would mean that AID works for this dataset. If some of the aspects in this inventory simulation are improved, we argue that the sequence should stay the same.}

\Changes{In case of the mixture approaches (this applies to ``Mixture'', ``Category Partial'' and ``Category Full''), the target service level was amended for each individual observation depending on the predicted probability of occurrence. This was done similar to how \cite{Svetunkov2023b} did it in Subsection 4.3 for the iETS model, using the formula:
\begin{equation} \label{eq:statCDFNew}
	F_z(z_{t+h} \leq Q) = \frac{F_y(y_{t+h} \leq Q) -(1 -\hat{p}_{t+h|t})}{\hat{p}_{t+h|t}},
\end{equation}
where $F_y(y_{t+h} \leq Q)$ is the target service level for the actual value $y_{t+h}$, $\hat{p}_{t+h|t}$ is the probability of occurrence conditional on the information on the observation $t$, $F_z(z_{t+h} \leq Q)$ is the confidence level calculated for the demand sizes $z_{t+h}$, and $Q$ is the quantile of interest.}

\Changes{We decided to focus on LightGBM because it produce the most accurate point forecasts in the previous subsection. However, we checked the inventory implications for other approaches and found similar results in terms of their ranking.}

%%%%%%%%%%%%%%% Inventory simulation results %%%%%%%%%%%%%%%
\subsection{\Changes{Inventory simulation results}}
\begin{figure}[htb!]
    \centering
    \includegraphics[width=\textwidth, page=1]{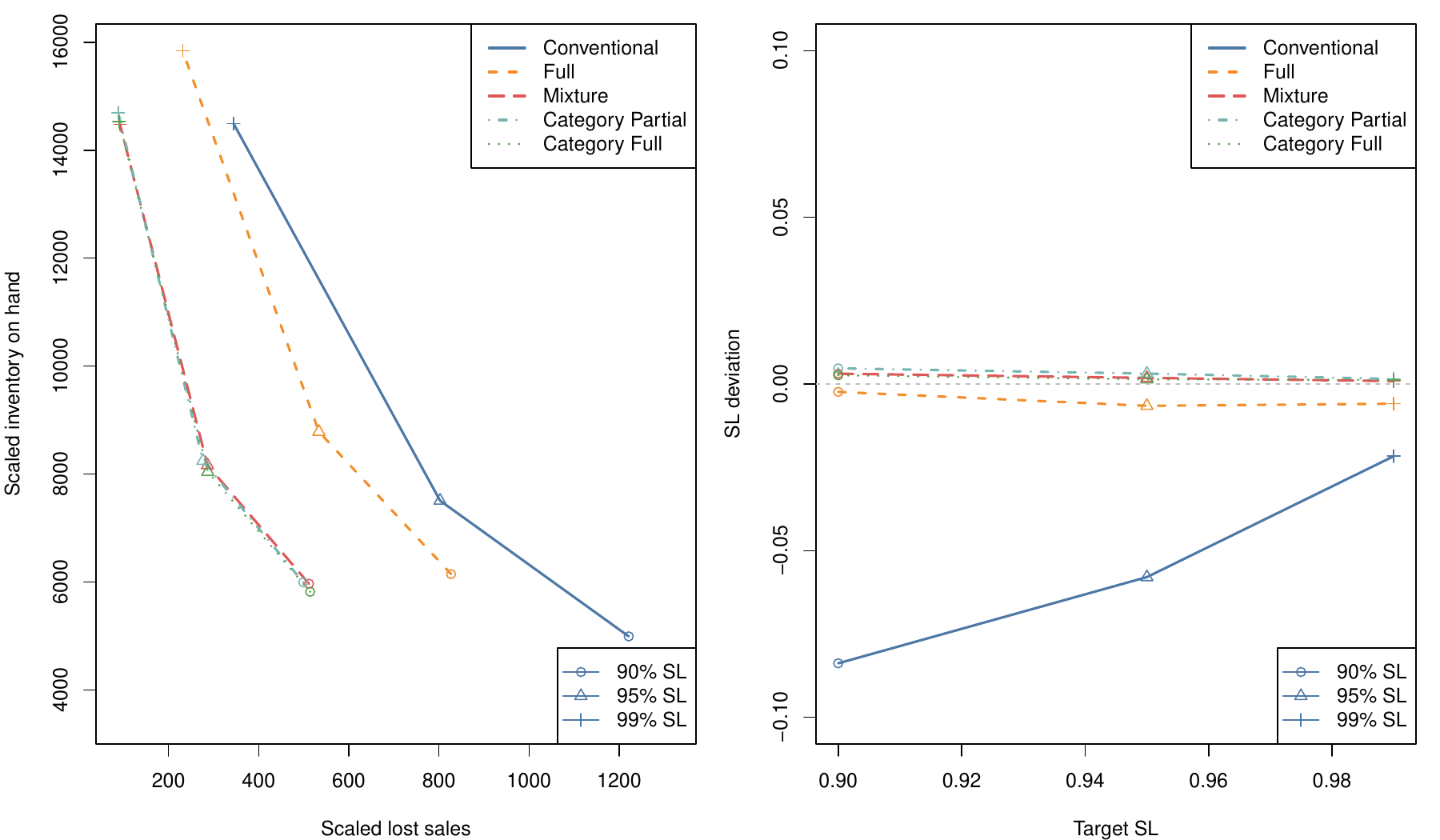}
    \caption{Lost sales vs inventory on hand and target service level vs the service level deviation. Origin 1.}
    \label{fig:Rewe-Inventory-Step1}
\end{figure}

\Changes{Lost sales and inventory on hand were scaled by dividing them by the in-sample standard deviation of the data. We also calculated ``Service level deviation'' by subtracting the achieved service level from the target one. Figure \ref{fig:Rewe-Inventory-Step1} shows these measures in one plot for the first step. For the lost sales vs inventory on hand image, the best performing approach is the one that is closest to the origin. For the service level image, the perfect approach should lie on the horizontal line going through zero (implying that the achieved service level equals the target one).}

\Changes{As we can see from the first image in Figure \ref{fig:Rewe-Inventory-Step1}, the ``Conventional'' approach leads to highest lost sales and inventory on hand in comparison with the other approaches. The ``Full'' approach which relies on stockouts detection improves upon it, leading to slightly higher inventory on hand but with the reduced lost sales. The categorisation approaches (which split the data into regular/intermittent and to regular/smooth/lumpy), together with the ``Mixture'' one bring further improvements, mainly reducing the lost sales. The three perform very similar, and their performance is indistinguishable. All three approaches produce the curve that is the closest to the origin, although it leads to slightly higher lost sales (at the cost of lower inventory on hand) for the low service level of 90\% than in case of the ``Full'' approach. When the service level is the highest (99\%), it seems to produce the best combination of the lost sales and inventory on hand.}

\begin{figure}[htb!]
    \centering
    \includegraphics[width=\textwidth, page=2]{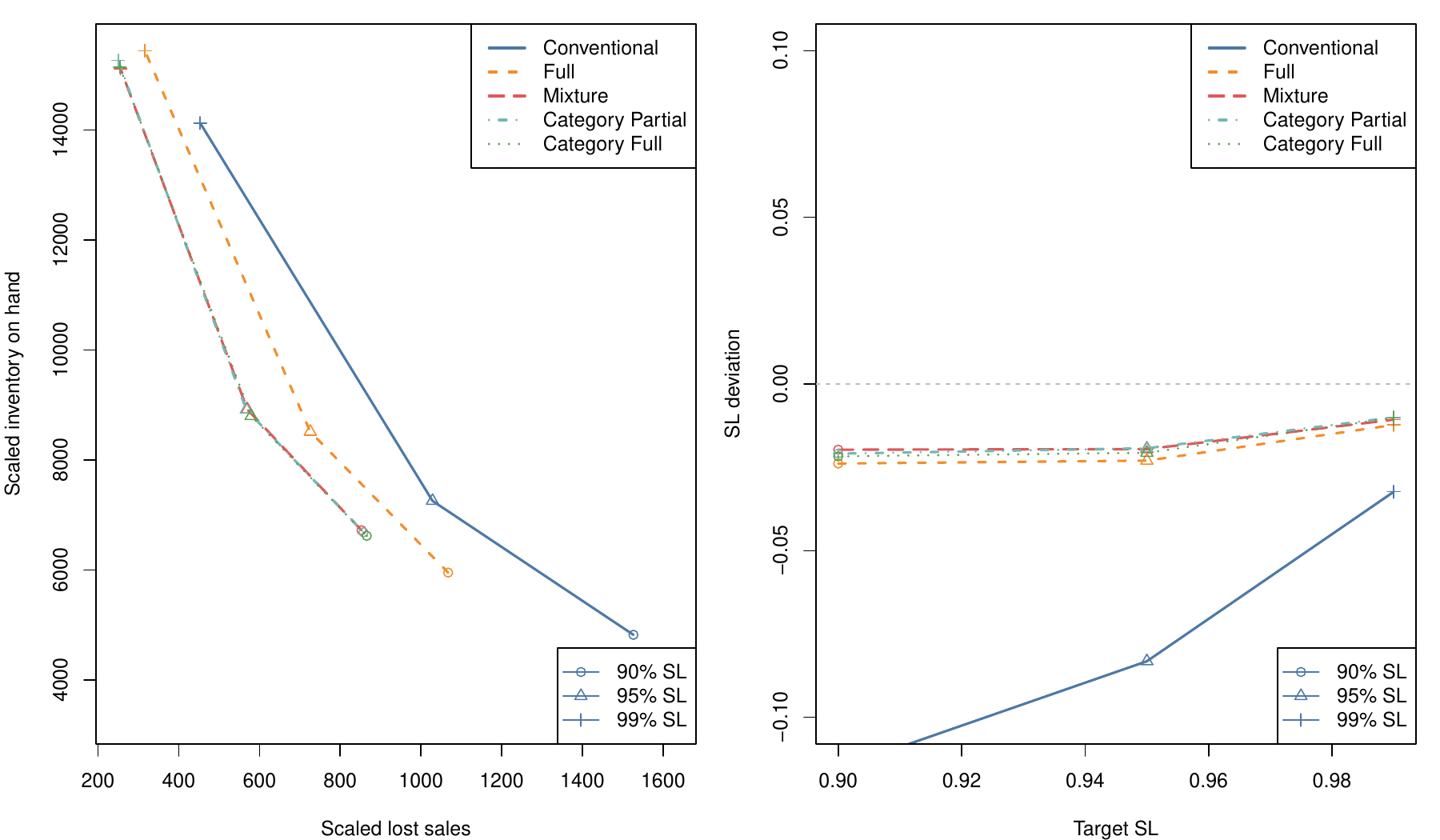}
    \caption{Lost sales vs inventory on hand and target service level vs the service level deviation. Origin 2.}
    \label{fig:Rewe-Inventory-Step2}
\end{figure}

\Changes{When it comes to the achieved vs target service level (the second image in Figure \ref{fig:Rewe-Inventory-Step1}), we see that the conventional approach consistently underperforms, containing fewer than expected observations. Moving to the ``Full'' approach, the difference becomes apparent: the improvement is substantial and the SL deviation becomes close to zero (being slightly below, implying that the the achieved SL is less than 1\% lower than the target). This once again demonstrates the importance of treating the stockouts correctly. Finally, moving to the ``Mixture'', ``Category Partial'' and ``Category Full'' approaches, the lines lie consistently slightly above zero, meaning that the achieved SL is higher than the target (the deviation is less than 1\%).}

\Changes{The whole situation changes for the second week as shown in Figure \ref{fig:Rewe-Inventory-Step2}, although the ranking stays the same, which shows that the AID principles work in terms of inventory management for the dataset. The three approaches (``Mixture'', ``Category Partial'' and ``Category Full'') still perform similar, leading to lower lost sales although at the cost of inventory on hand (in comparison with the other approaches). Furthermore, the three approaches lead to lower SL deviation than the other two approaches, although the difference between them and the ``Full'' one seems to be negligible.}

\Changes{As a summary from this experiment, we found that the inclusion of stockouts feature improves substantially the performance of LightGBM in terms of inventory. Furthermore, the split into demand occurrence and demand sizes leads to further improvements. However, the AID classification itself does not bring any benefits in terms of inventory performance in comparison with the ``Mixture'' approach.}

\Changes{Finally, we also included the count/fractional feature and experimented with different ways for generating quantiles based on this split (using different distributions and quantile generation approaches), but we did not find any improvement in terms of inventory performance in comparison with the ``Mixture'' approach. So, we argue that the split into count/fractional, while being theoretically appealing, is not practically useful.}

%%%%%%%%%%%%%%% Conclusions %%%%%%%%%%%%%%%
\section{Conclusions} \label{sec:Conclusions}
Intermittent time series are often met in a variety of contexts, including supply chain, retail etc. It is generally recognisable that intermittent demand should be treated differently than the regular one, yet it is not clear how to tell the difference between the two. In this paper, we discussed what intermittent demand is, focusing on why zeroes can happen in it. We argue that there are two fundamental reasons for them: (1) they can occur naturally because nobody buys a product at a certain point; (2) they can occur artificially due to disruptions or recording errors. We then moved to discussing possible types of demand, creating a classification based on the important fundamental demand characteristics, ending up with six categories, including regular/intermittent, intermittent smooth/lumpy and count/fractional ones.

After that, we developed an Automatic Identification of Demand (AID) approach that automatically detects stockouts and first classifies demand into regular or intermittent (depending on the presence of naturally occuring zeroes), and then does a fine split to one of the six categories based on AIC of several models underlying each of the types.

We tested the AID approach on simulated data to see how sensitive the stockouts detection part is and how accurate the demand classification one is. We found that the power of the stockouts detection mechanism is proportional to the sample size, probability of occurrence and the length of stockouts, being reverse proportional to the number of stockouts. The demand classification approach struggled in detecting the count data, in some cases flagging time series as fractional. This was because in some cases the models for fractional data can be efficiently used on the count one. We also found that its accuracy improves with the increase of the sample size.

Furthermore, we did an experiment on the real retail data, trying to see whether introducing specific features and using several fundamental modelling principles improves accuracy of several basic forecasting approaches. We found that:

\begin{itemize}
    \item Using a stockout dummy variable and capturing the level of data correctly (by removing the effect of stockouts) improves the accuracy of forecasting approaches;
    \item \Changes{The stockouts detection should be done for both the training and the test sets. If the series with stockouts are not removed from the test set, the forecasts would be evaluated incorrectly (i.e. showing which of the methods does better at predicting sales instead of demand);}
    \item Splitting the demand into demand sizes and demand occurrence, producing forecasts for each of the parts and then combining the result substantially improves the accuracy further;
    \item \Changes{Using the feature for regular/intermittent demand improves the forecasting accuracy, but does not seem to impact the inventory performance. Note that this separation is straightforward in AID: if after removing the stockouts, there are some zeroes left, the demand is identified as intermittent;}
    \item \Changes{The further split into smooth/lumpy leads to slight improvements in terms of accuracy, without a substantial impact on the inventory;}
    \item \Changes{The split into count/fractional demand does not bring value in terms of forecasting accuracy or inventory performance.}
\end{itemize}

We think that these findings have direct practical implications and can be used to improve accuracy of many forecasting approaches.

\Changes{Finally, we conducted additional experiment on the M5 daily data \citep{Makridakis2022}, setting the forecasting horizon to 14 days, applying the same approaches as in the case study to test the robustness of the AID approach. We found that the main principles discussed in this paper hold there as well.}

\Changes{In terms of the limitations, we note that other information criteria or cross validated error measures could be used instead of the AIC in the AID approach. We used it for simplicity, but an investigation of the impact of information criteria on the overall performance of the algorithm could be done in a future work.}

\Changes{We should also remark that we did not have the records of stockouts from the company to test whether the detection mechanism works well and in how many cases it makes a mistake. It would be interesting to explore this element in future work. But the more important message from the paper is that whatever detection mechanism you use for stockouts, they need to be treated and included in your model, because this becomes essential in terms of forecasting accuracy and inventory management.}

\section*{Data availability and competing interests}
The company data that support the findings of the case study are not available due to a non-disclosure agreement, the rest could be shared upon a request.

The authors declare no competing interests.

% \appendix
%%%%%%%%%%%%%%% Appendix %%%%%%%%%%%%%%%
% \section{Derivation of concentrated log-likelihood for the mixture model \eqref{eq:IG}} \label{sec:logLikDerivation}

\bibliographystyle{elsarticle-harv}
\bibliography{library}

\end{document}